\documentclass[10pt,twocolumn,letterpaper]{article}

\usepackage[pagenumbers]{cvpr} %
\usepackage{multirow}
\usepackage{longtable}
\usepackage{booktabs}
\usepackage{amsmath}
\usepackage{listings}
\usepackage{cuted}
\usepackage[T1]{fontenc}

\usepackage{tabularx}
\usepackage{tcolorbox}
\tcbuselibrary{listings, breakable, skins}
\usepackage{minted} %

\newtcolorbox{promptbox}{
  colback=gray!5!white,
  colframe=gray!50!black,
  fonttitle=\bfseries\footnotesize,
  title=Example JSON Prompt,
  breakable,
  enhanced,
  sharp corners,
  boxrule=0.5pt,
  left=4pt,
  right=4pt,
  top=4pt,
  bottom=4pt
}

\definecolor{cvprblue}{rgb}{0.21,0.49,0.74}
\usepackage[pagebackref,breaklinks,colorlinks,allcolors=cvprblue]{hyperref}

\def\eg{e.g\onedot}

\lstdefinelanguage{json}{
    basicstyle=\normalfont\ttfamily\footnotesize,
    morestring=[b]",
    stringstyle=\color{blue},
    morecomment=[l]{//},
    morecomment=[s]{/*}{*/},
    commentstyle=\color{dkgreen},
    morekeywords={true, false, null},
    keywordstyle=\color{mauve},
}

\def\paperID{} %
\def\confName{}
\def\confYear{}

\title{
Archon: A Unified Multimodal Model for Holistic Digital Human Generation
}

\author{
Chong Bao$^{1,2*}$\footnotemark[4]
\quad
Shichen Liu$^{2}$\footnotemark[1] \quad
Lijun Yu$^{3}$  \quad
David Futschik$^{2}$ \quad
Stylianos Moschoglou$^{2}$ \\
Shefali Srivastava$^{2}$ \quad
Ziqian Bai$^{2}$ \quad
Feitong Tan$^{2}$ \quad
Guofeng Zhang$^{1}$ \\
Zhaopeng Cui$^{1}$ \quad
Sean Fanello$^{2}$ \quad
Yinda Zhang$^{2}$\footnotemark[2] \\
$^{1}$State Key Lab of CAD\&CG, Zhejiang University \quad
$^{2}$Google \quad
$^{3}$Google DeepMind\\
}

\begin{document}

\twocolumn[{%
\renewcommand\twocolumn[1][]{#1}%
\maketitle
\begin{center}
    \centering
    \vspace{-1em}
    \captionsetup{type=figure}
    \includegraphics[width=1.0\linewidth, trim={0 0 0 0}, clip]{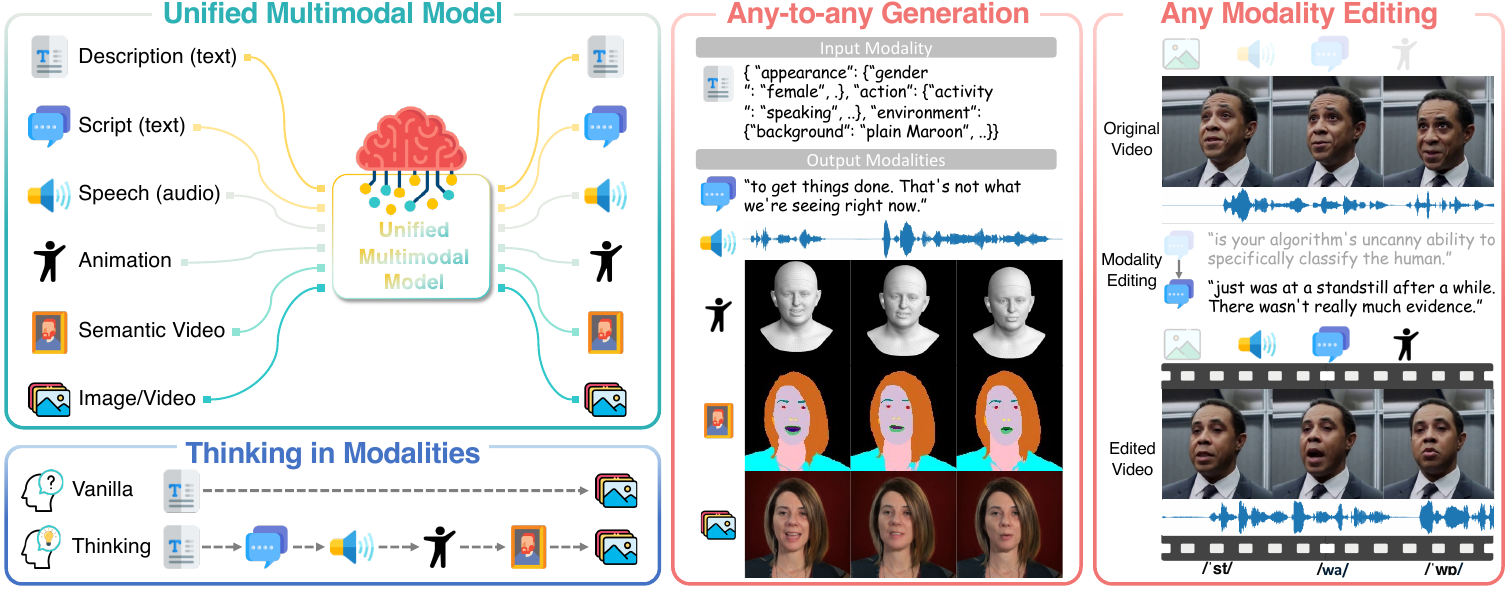}
    \captionof{figure}{\textbf{Archon}. We propose a novel unified multimodal model that performs cross-modal generation among a wide range of modalities, including description, script, speech, animation, semantic video, image, and video. Furthermore, we introduce the concept of \textit{Thinking in Modality} to reduce ambiguity during cross-modal transitions and enhance generation quality. Our model inherently supports conditional generation across arbitrary sets of modalities, enabling any modality editing throughout the entire multimodal input space.}
\label{Teaser}
\end{center}%
}]
\renewcommand{\thefootnote}{\fnsymbol{footnote}}

\footnotetext[1]{Authors contributed equally.}
\footnotetext[2]{Corresponding authors.}
\footnotetext[4]{The work was done during an internship at Google.}

\maketitle

\begin{abstract}
\vspace{-2em}

Digital humans are fundamental to immersive interaction, yet creating a unified model for holistic modalities, including text, audio, motion, and visual content, remains an open challenge. In this paper, we present Archon, a fully pretrained, human-centric unified multimodal model for holistic avatar generation. Archon unifies seven modalities with modality-specific tokenizers, and a native autoregressive unified multimodal model pretrained on synchronized modalities and 72 diverse tasks to model holistic joint distributions. To address the token explosion challenge in high-fidelity talking videos, we introduce a memory-efficient semantic video reparameterization achieving 4× token reduction while preserving fine-grained dynamics, coupled with a semantic-driven video diffusion decoder. We further propose a ``Thinking in Modality'' that decomposes ambiguous cross-modal tasks into stepwise thinking in an alternative chain of modality, progressively enhancing fidelity and controllability. Extensive experiments demonstrate that Archon achieves superior or comparable performance across diverse digital human generation tasks, validating the effectiveness of our unified framework. Project page:
\href{https://zju3dv.github.io/archon/}{https://zju3dv.github.io/archon/}.
\end{abstract}
    
\vspace{-1.8em}
\section{Introduction}
\label{sec:intro}
Digital humans are central to modern human–computer interaction, virtual reality, and digital entertainment. To support faithful representation, engaging communication, and flexible interaction, a unified model is required to seamlessly integrate a rich set of perceptual modalities, including text, audio, motion, semantic and color, capable of processing and generating each signal on demand. Despite decades of progress, creating fine-grained controllable digital humans that is able to process holistic multimodal signals remains a significant open challenge.

Current approaches for digital human systems are largely built upon specialized expert models that target individual modalities or sub-tasks, such as
speech-driven video generation~\cite{zhou2021pose, pang2023dpe, zhang2023sadtalker, xu2024vasa, cui2024hallo3, wei2025mocha, chen2025echomimic, tan2025fixtalk} and
image-conditioned text-to-speech~\cite{lee2023facetts, goto2020face2speech, ye2025emotional, kang2023facestylespeech}.
While these approaches achieve impressive realism within their respective domains, they share fundamental limitations: fragmentation and inefficiency. Training on distinct, modality-specific datasets introduces distribution mismatches, making the coordination of different expert models as a unified system brittle for novel multimodal tasks. This fragmentation also leads to redundant capacity and limited scalability, as each expert model must independently learn task-specific knowledge that could otherwise be shared in a modality-driven unified space. Crucially, the introduction of any new modality demands training a new model from scratch or non-trivial finetuning of a model designed for a different task. In contrast, we advocate for a holistic, unified, \textit{any-to-any} multi-modal generative framework that overcomes these issues by reusing shared representations and seamlessly adapting to novel tasks without requiring dedicated model pretraining.

Existing unified multimodal models are not truly “holistic” in the context of digital humans. While current multimodal language models (MLLM)~\cite{alayrac2022flamingo, chowdhery2023palm, team2023gemini, peng2023kosmos, Qwen3-Omni, Qwen3-VL, lu2024deepseek} focus on understanding multimodal inputs, their outputs are confined to text. Conversely, contemporary unified generative models can generate text, image or video but overlook the supports on audio ~\cite{xie2025show, wu2024next, zhan2024anygpt, deng2025emerging, zhou2024transfusion}, or generate only non-speech audio such as music or environmental sounds~\cite{wu2024next, zhan2024anygpt, kondratyuk2023videopoet}. Specifically for digital human, parsing human speech, animation (\eg, through 3D Morphable Model), and preserving identity across the temporal dimension are barely studied in the domain of multimodal models.

To address these challenges, we introduce Archon, a  pretrained unified multimodal model designed for holistic digital human generation. Archon pushes general-purpose multimodal reasoning to domain-specific avatar synthesis, enabling holistic cross-modal generation and understanding across a comprehensive set of modalities, including description, speech (transcript and audio), animation, semantic video, image, and video.
First, we design a suite of modality-specific tokenizers that encode heterogeneous signals from each modality into discrete integer tokens, which are subsequently merged into a unified vocabulary. These tokenizers are crafted to balance reconstruction fidelity with token-sequence efficiency, facilitating compact yet expressive representations across modalities.
Second, we propose a native unified multimodal model that models the joint distribution of holistic modalities in a autoregressive framework. To support large-scale pretraining, we curate a comprehensive multimodal dataset comprising synchronized modality pairs and 72 diverse multimodal tasks, enabling the model to learn rich cross-modal correspondences.

Third, human-centric talking videos inherently involve high-frequency dynamics, such as lip motion, facial expressions, and head poses, which demand high frame rates (e.g., 30 fps). However, a 5-second 30 fps video at 256×256 resolution amounts to 9K tokens~\cite{yu2024language}, exceeding the context window of MLLMs (8K tokens)~\cite{anil2023palm2}.
Moreover,
MLLM's reasoning on discrete tokens disrupts the continuous nature of video signals, thereby degrading generation quality. To overcome this, we introduce a memory-efficient video discretization tailored for cross-modal reasoning, coupled with a semantic-driven video diffusion decoder for high-fidelity video synthesis. Specifically, we replace RGB videos with semantic videos, which consist of discrete semantic labels and preserve essential dynamics and structure while discarding redundant visual information, achieving a 4× reduction in token count. Subsequently, a video diffusion model conditioned on semantic video, reference images, and textual descriptions generates high-quality videos.

Fourth, certain cross-modal generation tasks suffer from substantial domain gaps and inherent ambiguities. For instance, generating video from speech necessitates both extracting explicit information (e.g., gender) and synthesizing missing details (e.g., appearance, expression), which challenges the scaling-to-quality rule of expert models. Leveraging the flexibility of our unified model in holistic cross-modal reasoning, we introduce a ``Thinking in Modality'' strategy during inference. This approach decomposes complex cross-modal tasks into stepwise thinking in an alternative chain of modality generation, progressively enhancing fidelity and controllability in cross-modal synthesis.

\noindent Our contributions can be summarized as:
\begin{itemize}
    \item We introduce Archon, a fully pretrained, human-centric unified multimodal model with holistic avatar synthesis via a native MLLM pretrained on synchronized modalities and 72 tasks to model holistic joint distributions.
    \item We propose a memory-efficient video discretization and a semantic-driven diffusion decoder for efficient and high-quality generation, and a ``Thinking in Modality'' inference strategy to enhance fidelity and controllability.
    \item Extensive experiments demonstrate superior or comparable performance with expert models across holistic digital human generation tasks, validating the effectiveness of our unified framework.
\end{itemize}

\section{Related Works}
\label{sec:related}

\begin{figure*}[ht!]
    \centering
    \includegraphics[width=\textwidth]{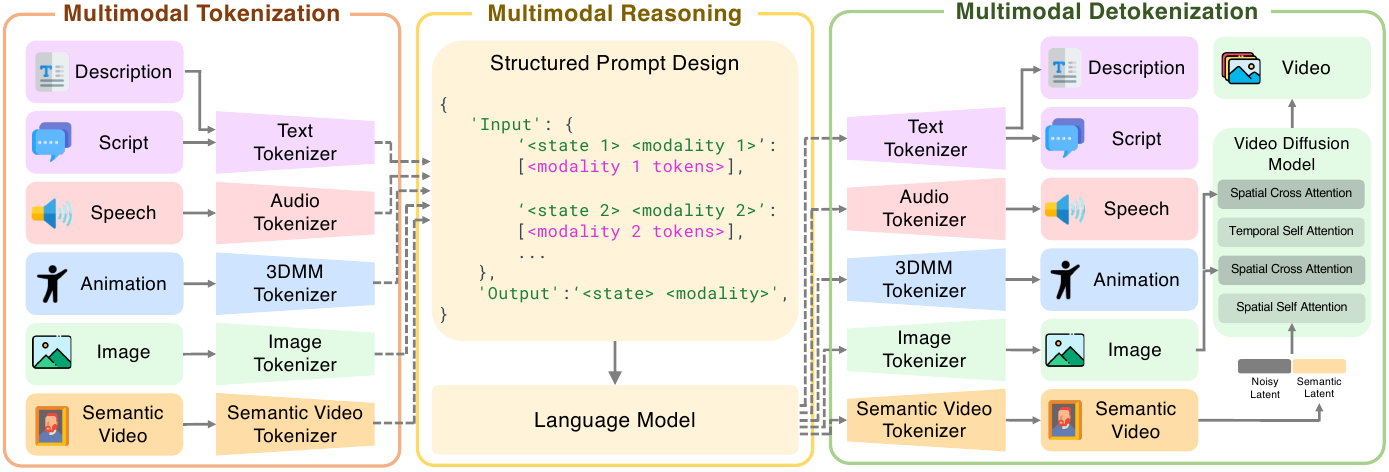}
    \caption{\textbf{Pipeline}. We use modality tokenizers to tokenize description, script, speech, animation, image and semantic video into discrete tokens. These tokens are arranged in a structured format and input to the language model for multimodal reasoning. The synthesized tokens are detokenized into raw modalities. For synthesizing high-quality video, we employ a semantic-driven video diffusion model to synthesize high-quality video conditioned on image and semantic segmentations.}
    \label{fig:pipeline}
\end{figure*}

\noindent\textbf{Digital human generation.}
Research on digital humans has evolved from modality-specific modeling towards increasingly unified representations.
Early studies focused on specialized tasks,
such as facial modeling and reconstruction from images~\cite{feng2021deca, danvevcek2022emoca},
talking heads generation and lip synchronization from audio~\cite{cui2024hallo3, chen2025echomimic, zhang2023sadtalker, wei2025mocha, wei2024aniportrait},
personalized speech generation from identity~\cite{lee2023facetts},
and similar isolated tasks.
While these expert systems achieved high fidelity within their domains, they typically rely on task-specific model architecture, limiting cross-modal generalization.
To overcome these constraints, more recent efforts aim to couple multiple tasks or modalities into unified human-centric frameworks.
Human foundation models such as Sapiens~\cite{khirodkar2024sapiens} and OmniHuman~\cite{lin2025omnihuman} move toward large-scale representations pretraining that jointly capture pose, geometry, and semantic understanding, while promptHMR~\cite{wang2025prompthmr} introduces language-guided controllability in mesh recovery.
Despite advances, most existing systems still treat inputs or outputs as modality-specific, rather than learning a truly unified interface across modalities. Consequently, these systems struggle with any-to-any modality translation.

\noindent\textbf{Multimodal language models.}
In parallel, multimodal models~\cite{wu2024next, zhan2024anygpt, kondratyuk2023videopoet, deng2025emerging} demonstrate that diverse data types can be represented within a unified transformer backbone.
Models such as Flamingo~\cite{alayrac2022flamingo}, PaLM-E~\cite{driess2023palm}, and Kosmos~\cite{peng2023kosmos} extend language models with multimodal context, improving cross-modal reasoning and conditional generation.
The development of modality-specific discrete tokenizers~\cite{2020t5, zeghidour2021soundstream, yu2023magvit, yu2024language} further enables unified sequence modeling across text, audio, video, and other signals. This paradigm is explored by AudioLM~\cite{borsos2023audiolm}, SpeechGPT~\cite{zhang2023speechgpt}, VideoPoet~\cite{kondratyuk2023videopoet}, and large unified architectures such as Gemini~\cite{team2023gemini}, showing that a single transformer can perform both perception and generation through a shared token space.
Building on these insights, we extends unified multimodal model into the domain of avatar.

\section{Method}

We propose a unified multimodal model for holistic digital human generation and understanding.
As shown in Fig.~\ref{fig:pipeline}, our model consists of four core modules: (1) a set of modality-specific tokenizers (Sec.~\ref{sec:tokenizer}), (2) a language model backbone for cross-modal reasoning (Sec.~\ref{sec:llm}), (3) a semantic-driven diffusion model for high-quality video generation (Sec.~\ref{sec:diffusion}), (4) ``Thinking in Modality'' strategy for high quality and controllability generation (Sec.~\ref{ssec:thinking}).

\subsection{Multimodal Tokenization}
\label{sec:tokenizer}
As shown in Fig.~\ref{fig:pipeline}, we model a comprehensive coverage of modalities in the digital human, which are \textit{description}, \textit{script}, \textit{speech}, \textit{animation}, \textit{semantic video}, \textit{image} and \textit{video}. In the following paragraphs, we introduce the tokenization process of each modality. The details of the model architectures and training strategies are in supplementary Sec.~\textcolor{cvprblue}{D}.

\noindent\textbf{Image tokenizer.}
We employ the pretrained MAGVIT-v2~\cite{yu2024language} model for image tokenization, selected for its superior visual fidelity and compression capabilities.
MAGVIT-v2~\cite{yu2024language} is a 3D convolution-based VQGAN~\cite{esser2021taming} that utilizes a lookup-free vocabulary with a codebook size of $2^{18}$. We leverage this model to quantize images of $256\times256$ resolution into a discrete $16\times16$ token representation.

\noindent\textbf{Semantic video as context-efficient video tokenizer.}
Directly adopting existing video tokenizers~\cite{yu2024language} is impractical in multimodal model, as their token output often exceeds the context window capacity of language models.
As a case in point, a 32GB-HBM TPUv6 can accommodate a context window of approximately 8K tokens. However, encoding a five-second, 256$\times$256 video (30 fps) produces 9K tokens~\cite{yu2024language}.
Furthermore, the sheer magnitude of video tokens introduces a severe modality imbalance. A five-second audio clip, for instance, generates only 940 tokens, representing a small fraction of the video token count. This discrepancy leads to a training bias, as video tokens overwhelmingly dominate the dataset.
Na\"ive token reduction via higher-compression tokenizers is infeasible, as this necessitates retraining a new tokenizer with an intractably large vocabulary (exceeding $2^{18}$) and network bottleneck, which in turn significantly increases the language model's modality learning complexity.

Inspired by semantic-to-video generation works~\cite{mallya2020world}, we introduce a novel, memory-efficient discretization strategy that decomposes the video into a reference image and a semantic video. The reference image is typically the first video frame.
The semantic video comprises 21 discrete semantic categories (e.g., eyelid, eyebrow, nose) obtained from an off-the-shelf face segmentation model~\cite{oquab2023dinov2}.
The semantic video retains crucial structural and motion information while discarding texture, significantly reducing the information density but preserving the pixel-domain nature of the video. The discrete nature of the semantic labels induces a smooth distribution along the spatial dimension and aligns naturally with the autoregressive reasoning paradigm of large language models, which allows a low-resolution semantic video to effectively represent a high-resolution video. A video diffusion model is subsequently used to upsample and generate a high-quality video conditioned on the semantic video and reference image (detailed in Sec.~\ref{sec:diffusion}).

Specifically, we learn a semantic video tokenizer to compress a $L\times128\times128$ semantic video into $(\frac{L-1}{4}+1) \times 8 \times 8$ tokens, where $L$ is the number of frames. We finetune MAGVIT-v2~\cite{yu2024language} to reuse its powerful pretrained encoders. An \textit{invertible color embedding} is defined to map each semantic label to a distinct RGB color, allowing the semantic video to be processed as a standard color video.
To align with the reduced complexity of this representation, the codebook size of our semantic tokenizer is set to $2^{10}$.

\noindent\textbf{Speech tokenizer.}
We use a pretrained SoundStream tokenizer~\cite{zeghidour2021soundstream} for speech tokenization.
SoundStream~\cite{zeghidour2021soundstream} employs a residual vector quantizer (RVQ) that encodes 16\,kHz audio into discrete tokens at 25\,frames per second across 8 residual levels.
Following the VideoPoet~\cite{kondratyuk2023videopoet}, we retain only the first 4 RVQ levels for efficiency and stability. Each of these residual levels has an independent vocabulary of 1,024 codes. The final token sequence is constructed by arranging tokens from lower to higher levels, preserving the hierarchical structure of the residual quantization.

\noindent\textbf{Animation tokenizer.}
We utilize 3D Morphable Model (3DMM) parameters~\cite{flame} to parameterize the animation of digital humans. The 3DMM have three components: shape, expression, and pose. The shape parameters capture the subject’s static geometry and appearance, remaining invariant throughout the video clip, and are represented by an $e_{sh}$-dimensional vector. In contrast, the expression and pose parameters evolve temporally and are modeled as continuous 1D temporal signals, with respective dimensions of $L\times e_{exp}$ and $L\times e_{pose}$. We independently learn three Residual Vector Quantized VAE (VQVAE) models~\cite{van2017neural} for shape, expression and pose. Each employs causal convolution to discretize parameters. The resulting codebooks are configured as follows: shape (8 levels, 512 codes per level), expression (8 levels, 2048 codes per level), and pose (6 levels, 512 codes per level). Consistent with the speech tokens, the final RVQ token sequence is constructed by arranging tokens from lower to higher residual levels.

\noindent\textbf{Text tokenizer.}
To preserve the inherent text generation capabilities of our language model, we retain its original tokenizer, the T5 encoder~\cite{raffel2020exploring}, for all text processing tasks.

\subsection{Language Model for Multimodal Generation}
\label{sec:llm}

We employ an auto-regressive language model that supports flexible generation by conditioning on prompts contextualized with task and modality-specific token cues.

\noindent\textbf{Task definition.}
Different from typical multimodal settings that focus on only a few generation tasks, the input and output of the holistic digital human generation problem can be arbitrary modality sets.
We categorize the modalities of interest into
\textit{time-dependent} ones $D_{\mathrm{var}} = \left\{ d_\mathrm{scr}, d_\mathrm{spc}, d_\mathrm{exp}, d_\mathrm{sem}, d_\mathrm{vid} \right\}$, including \textit{script}, \textit{speech}, \textit{semantic masks}, and \textit{video};
and \textit{time-invariant} ones $D_{\mathrm{inv}} = \left\{ d_\mathrm{img}, d_\mathrm{dsc}, d_\mathrm{shp} \right\}$, including \textit{image}, \textit{description}, and \textit{shape}.
We define the full modality set as
\begin{equation}
    \mathcal{D} =  \{\, d^t \mid d \in D_{\mathrm{var}}^{t},\; t \in \{\mathrm{0}, \mathrm{1}\} \,\}  \;\cup\; D_{\mathrm{inv}},
\end{equation}
where $d^0, d^1$ denote the modality in the past and present tense respectively.
Given input modalities $\mathcal{D}_{\mathrm{cond}} \subset \mathcal{D}$ as conditions, the goal of a generation task is to generate a subset of the remaining modalities $T:\mathcal{D}_\mathrm{cond} \rightarrow \mathcal{D}_\mathrm{gen}$, where $\mathcal{D}_{\mathrm{gen}} = \{\, d_j \mid d_j \notin \mathcal{D}_{\mathrm{cond}} \,\}$.
Existing works~\cite{kondratyuk2023videopoet, wang2023visionllm} employ a special task token to instruct the task type.
However, as the number of modalities increases, the possible combinations of input and output causes combinatorial explosion, making it increasingly difficult for the model to handle all tasks within a unified framework.
To address this challenge, we reformulate the auto-regressive generation process allowing an arbitrary combination as
\begin{equation}
T_j:
\begin{cases}
\mathcal{D}_{\mathrm{cond}} \;\rightarrow\; d_1, & j = 1, \\
\mathcal{D}_{\mathrm{cond}} \cup \{ d_1, \ldots, d_{j-1} \} 
    \;\rightarrow\; d_j, & j > 1.
\end{cases}
\label{eq:autoregressive}
\end{equation}
This recurrent formulation makes explicit that, at each step \(j\), the model generates one modality \(d_j\) conditioned on the input modalities along with all previously generated ones.
Consequently, rather than training the model to generate all modalities jointly, we train it to predict \textit{one modality at a time}.
This reduction preserves the expressive power of the original formulation while substantially simplifying the task space.
For clarity, we henceforth define sequential generation order from $d_1$ to $d_{\vert \mathcal{D}_{\mathrm{gen}} \vert}$ as $\mathcal{D}_{\mathrm{cond}} \rightarrow [ d_1, \dots, d_{\vert \mathcal{D}_{\mathrm{gen}} \vert} ]$.

\noindent\textbf{Prompt design.}
To instruct the model of input and desired output,
the input prompt, which is a series of tokens, must contain necessary task information.
Existing language models~\cite{comanici2025gemini, chowdhery2023palm, anil2023palm2, grattafiori2024llama} are good at parsing and understanding the structured inputs, e.g., HTML, JSON, etc.
Rather than employing special tokens to denote modalities or tasks, we organize the prompt as a structured data serialization filled with natural language descriptors.
These descriptors serve as keys that explicitly delineate the modality type, state, input, and expected output, as illustrated in Fig.~\ref{fig:pipeline}.
This design has two advantages: (1) it mitigates the burden of the model that heavily relies on a sparse set of special tokens; (2) it provides clearer semantic grounding that can leverage the knowledge in the pretrained language model.

\noindent\textbf{Language model backbone.}
We adopt PaLM2~\cite{anil2023palm2}, a prefix decoder-only model with bidirectional prefix attention. 
The model processes an input sequence of tokens and autoregressively predicts the subsequent token from a unified vocabulary. This vocabulary partitions different modalities into distinct contiguous index ranges; for instance, text tokens occupy indices 0–256127, while video tokens are assigned to 256128–518272. Each token is associated with its own learnable embedding vector.

\noindent\textbf{Multimodal training strategy.}
To equip our model with the ability to map across arbitrary modality sets, we conduct training across 72 diverse multimodal tasks and enable compositional inference (see supplementary Sec.~\textcolor{cvprblue}{C.3} for details).
Following the AGD training paradigm~\cite{akbari2023agd}, we employ a context window of 8K tokens, dynamically populated at each training step by sampling from our task distribution.
However, a na\"ive random sampling approach, selecting one task per step, is susceptible to three forms of training bias: (1) Model bias, where gradients backpropagate towards a sub-optimal direction for a single-modality task rather than the global gradient for multiple modality tasks. (2) Distributional bias, where random sampling over an uneven task distribution causes the model to under-prioritize modalities with fewer tasks. (3) Difficulty variance, where tasks of greater difficulty intrinsically require more learning effort.
To overcome these challenges, we first sample multiple tasks at each step, ensuring the model learns the joint distribution of multiple modalities. Second, we propose a novel sampling strategy to balance the multimodal training, factoring in both the task count per modality and the task difficulty.
The sampling weight $S(i)$ for the $i$-th task is defined as:
$S(i) = \frac{\log(p_i)}{N_{m(i)}}$
where $p_i$ is the perplexity of the $i$-th task, and $N_{m(i)}$ is the total number of tasks with the output modality being  $m(i)$. The perplexity $p_i$ measures task difficulty and estimated using a baseline model trained with uniform sampling weights.
This strategy balances model capacity to learn joint distribution across all tasks.

\begin{figure}[t!]
    \centering
    \includegraphics[width=1.0\linewidth, trim={0 0 0 0}, clip]{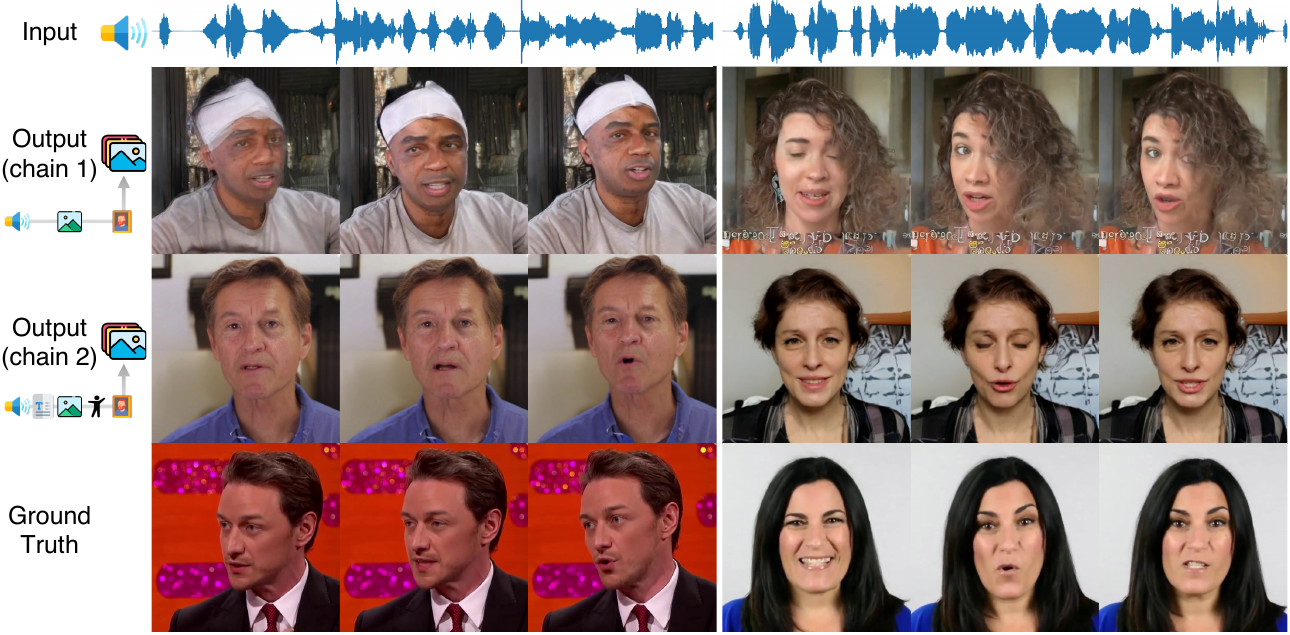}
    \vspace{-2em}
    \caption{\textbf{Thinking in Modality}. We show the results of speech-to-video generation with different thinking strategies. The videos generated from chain 2 contain less distortion (e.g., blurry appearance and undefined textual symbols) and exhibit closer identity alignment with the ground truth.}
    \vspace{-1em}
    \label{fig:ablation}
\end{figure}

\vspace{-0.5em}
\subsection{Semantic-driven Video Diffusion Model}
\label{sec:diffusion}
While the language model leverages discrete semantic videos as a memory-efficient alternative to continuous RGB videos for cross-modal reasoning, we introduce a semantic-driven video diffusion model that decodes these semantic videos into high-fidelity outputs, guided by a reference image and textual description.

\noindent\textbf{Semantic and appearance conditioning.}
We employ WALT~\cite{walt}, a transformer-based latent diffusion model originally designed for text-to-video generation, as our video diffusion backbone. We modify its architecture to accept low-resolution reference images, semantic masks, and textual descriptions as inputs for high-quality video generation. To provide strong motion guidance, we concatenate the encoded semantic mask latents with the noisy video latents along the feature dimension before feeding them into the diffusion model.
We utilize a cross-attention module to condition the reference image. To enhance the correspondence between video latents and the reference image, we extract the semantic mask of the reference image and concatenate its encoded semantic latent with the reference image latent along the feature dimension, analogous to our semantic signal conditioning. This composite feature is then concatenated with text embeddings to perform cross-attention with the video latent. In this way, semantics serves as a structural bridge, facilitating the effective transmission of the reference appearance into the generated video.

\noindent\textbf{Diffusion training.}
We fine-tune the entire WALT~\cite{walt} backbone to adapt it for our semantic-driven video generation task. The input reference images and semantic masks are processed at $256\times256$ resolution, while the output videos are synthesized at $512 \times 512$ resolution. The model employs a $\mathbf{v}$-prediction parameterization~\cite{salimans2022progressive} and is optimized using Mean Squared Error (MSE) loss between the predicted and target velocities.

\begin{figure*}[ht!]
    \centering
    \includegraphics[width=0.97\linewidth, trim={0 0 0 0}, clip]{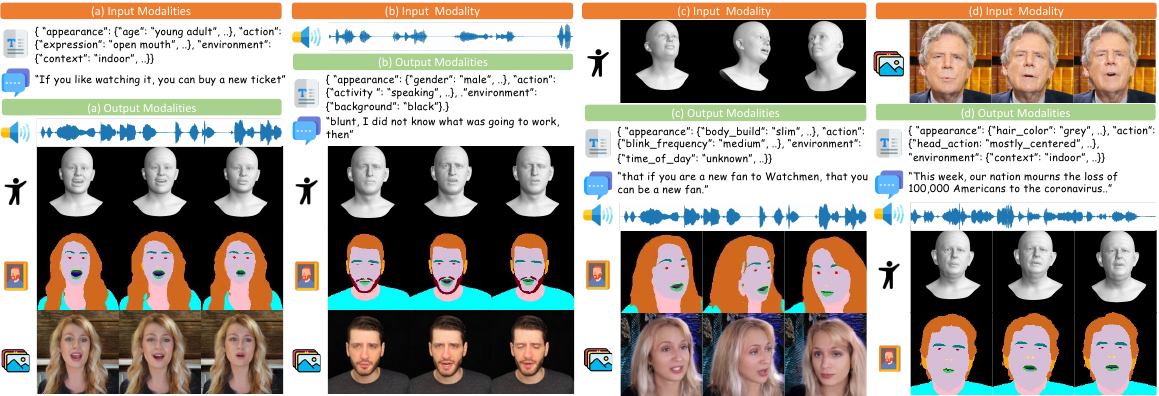}
    \vspace{-0.5em}
    \caption{\textbf{Multimodal Generation.} We show holistic modality generation and understanding given an arbitrary modality as input.}
    \label{fig:video_gen}
\end{figure*}

\subsection{Thinking in Modality}
\label{ssec:thinking}

While our model supports direct generation between modalities, we observe that different modality transitions exhibit varying levels of uncertainty.
For instance, generating video directly from speech typically shows higher perplexity than from 3DMM shape and expression, as shown in Fig.~\ref{fig:ablation}, indicating that the latter offers a more controllable and interpretable pathway.
Motivated by this, we introduce \textit{thinking in modality} --- an inference strategy that explores alternative generative chains by instructing the model to produce intermediate modalities that form a smooth transition in semantic granularity, thereby reducing uncertainty and improving output quality.
This approach requires no retraining and simply exploits the model’s ability to condition on multiple modalities in a unified framework.

\section{Experiments}

\subsection{Experiment Setup}

\noindent\textbf{Models.}
We fine-tune a pretrained 1B PaLM2~\cite{anil2023palm2} featuring a 550K vocabulary to equip the model with the language priors.
The vocabulary size of language model is 550K for multimodal tokens.
We employed the Adam optimizer~\cite{adam} with a weight decay of $10^{-3}$ and a 4,000-step learning rate warmup from $0$ to $10^{-3}$, followed by a cosine decay schedule to $5 \times 10^{-5}$. We use 256 TPUv6 Trillium to train the language model for 20 days with a batch size of 256.
For the diffusion model, we employ Adafactor optimizer~\cite{shazeer2018adafactor} and a 6,000-step linear warmup from $0$ to $1.2 \times 10^{-4}$ followed by a constant schedule. We use 128 TPUv6 to train the diffusion model for 10 days with batch size 128.

\begin{figure*}[ht!]
    \centering
    \includegraphics[width=1.0\linewidth, trim={0 0 0 0}, clip]{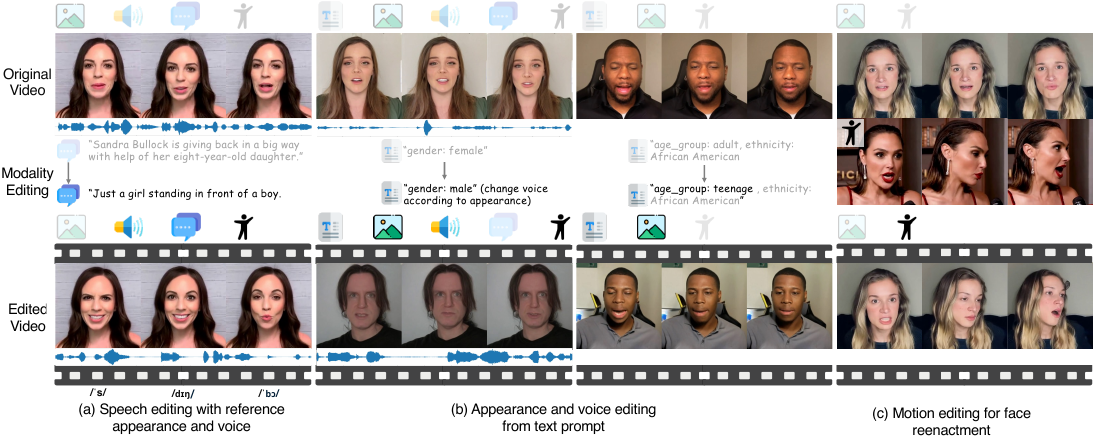}
    \vspace{-1.5em}
    \caption{\textbf{Modality-specific Editing}. We show flexible any modality editing to modify an arbitrarily chosen modality while maintain the others untouched. The icons on the top show the modalities used in the example, and the highlighted icons are the ones that are edited.}
    \label{fig:editing}
\end{figure*}

\noindent\textbf{Datasets.}
We train our model on a comprehensive multimodal dataset with 6,000 hours of monologue videos sourced from the public Internet.
Each video contains speech and script synchronized to the video~\cite{syncnet}.
We caption the video with Gemini 2.5 Pro~\cite{comanici2025gemini}, fit the 3DMM parameters following~\cite{wood2021fake}, and extract the face segmentation using our trained segmentation models with a DinoV2 backbone~\cite{oquab2023dinov2}.
We evaluate our method on two standard benchmarks: CelebV-HQ~\cite{zhu2022celebv} and HDTF~\cite{hdtf} that are disjoint from our training dataset. From each, we randomly sample a test set of 200 videos following previous work~\cite{chen2025echomimic, cui2024hallo3}, and extract the script from the speech using Whisper~\cite{radford2023whisper}.

\subsection{Multimodal Generation and Editing}
\noindent\textbf{Multimodal generation.}
We demonstrate our model's any-to-any generation capabilities in Fig.~\ref{fig:video_gen}. First, we showcase text-to-all generation. Given a description of a person's appearance, action, and environment, optionally augmented with a script, our model sequentially generates all corresponding modalities: script, speech, 3DMM, segmentation, and the talking video. As shown in Fig.~\ref{fig:video_gen}(a), our model generates realistic talking videos of a woman, with speech and visuals that precisely adhere to the input attributes.
Second, for speech-to-all generation, our model infers the person’s appearance (as a description) and script from the input speech, subsequently using these to generate the talking video, 3DMM, and segmentation. As illustrated in Fig.~\ref{fig:video_gen}(b), the model correctly interprets the person’s gender from the audio and generates a video that is temporally synchronized with the input speech.
Third, our model supports more fine-grained control using 3DMM or segmentation as inputs. When conditioned on these signals, it convincingly generates high-fidelity videos with responsive head motion and accurate expressions that match the input, as shown in Fig.~\ref{fig:video_gen}(c).
Finally, our model exhibits a strong capacity for understanding high-level modalities from low-level video, semantic, and 3DMM inputs. As shown in Fig.~\ref{fig:video_gen}(d), our model extract a description of the person's appearance, action, and environment; read the script directly from lip motion (lip-reading); synthesize a semantically-coherent voice for the input signals; reconstruct the 3DMM coefficients and semantic segmentations from the input video.
Please see supplementary Sec.~\textcolor{cvprblue}{E.1} for more examples.

\noindent\textbf{Modality-specific editing.}
We demonstrate the modality-specific editing capability of our model in Fig.~\ref{fig:editing}.
Firstly, we showcase script editing in Fig.~\ref{fig:editing}(a).
Our model synthesizes new speech following the edited script in the original voice and new video with synchronized lip motion and animation while maintain the identity.
Secondly, our model can seamlessly change the appearance of the subject via text prompt.
As shown in Fig.~\ref{fig:editing}(b), changing the gender or appearance attributes in the text description synthesizes a new talking video, retaining all other original aspects.
The speech is optionally modified to match the new appearance while still following the original script.
Finally, we perform face reenactment by editing the head motion using a reference video, as shown in Fig.~\ref{fig:editing}(c).
Our model generates high-quality videos with animation aligned with the reference, even under extreme head poses.

\subsection{Comparisons on Modality-Specific Tasks}

Although our model is built for broad multimodal generation, we investigate its performance on canonical modality-specific benchmarks to assess the unified architecture’s capability in handling individual tasks.
Note that our multimodal model is not trained on the benchmark dataset and is used straightforward on these tasks without any finetuning.

\begin{table}[!t]
    \centering
    \scriptsize
\tabcolsep 7.1pt
\begin{tabularx}{\linewidth}{lccccc}
    \toprule
    \multirow{2}{*}{Methods} & \multicolumn{5}{c}{CelebV-HQ~\cite{zhu2022celebv}} \\
     \cmidrule(l){2-6}
        ~ & FID↓ & FVD↓ & Sync-C↑ & Sync-D↓ & IQA↑  \\  
        \midrule
        AniPortrait$^{*}$~\cite{wei2024aniportrait} & 39.73 & 160.7 & 3.493 & 10.982 & \textbf{3.833}  \\ 
        EchoMimic$^{*}$~\cite{chen2025echomimic} & 56.81 & 236.9 & 4.463 & 9.575 & 3.601 \\
        Hallo3~\cite{cui2024hallo3} & \underline{15.67} & \underline{105.5} & \textbf{5.429} & \underline{9.158} & 3.722  \\ 
        Ours & \textbf{6.818 }& \textbf{93.81 }& \underline{5.210} & \textbf{8.998} & \underline{3.794} \\
    \midrule
    \multirow{2}{*}{Methods} & \multicolumn{5}{c}{HDTF~\cite{hdtf}} \\
         \cmidrule(l){2-6}
    ~ & FID↓ & FVD↓ & Sync-C↑ & Sync-D↓ & IQA↑ \\
    \midrule
    AniPortrait~\cite{wei2024aniportrait} & 42.03 & 162.8 & 2.879 & 10.889 & 3.813 \\
    EchoMimic$^{*}$~\cite{chen2025echomimic} & 45.90 & 241.6 & 5.467 & 9.36 & 3.743  \\
    Hallo3$^{*}$~\cite{cui2024hallo3} & \underline{12.78} & \underline{96.51} & \textbf{6.376} & \underline{9.131} & \underline{3.83}  \\
    Ours & \textbf{5.779} & \textbf{81.64} & \underline{6.198} & \textbf{8.822} &\textbf{ 3.94} \\
  \bottomrule
\end{tabularx}
\vspace{-1.0em}
    \caption{
    \textbf{Comparisons on speech-driven video generation.} We compare video generation quality and lip synchronization against baselines. ``$*$'' denotes methods trained on the benchmark dataset.
    }
    \vspace{-1.5em}
    \label{tab:audio_driven_gen}
\end{table}

\begin{table}[!t]
    \centering
\tabcolsep 2.5pt
\scriptsize
\begin{tabularx}{\linewidth}{lccc ccc}
    \toprule
    \multirow{2}{*}{Methods} & \multicolumn{3}{c}{CelebV-HQ~\cite{zhu2022celebv}} & \multicolumn{3}{c}{HDTF~\cite{hdtf}}\\
     \cmidrule(l){2-4}
     \cmidrule(l){5-7}
        ~ & MCD-DTW↓ & C-SIM↑ & Id. Acc.↑ & MCD-DTW↓ & C-SIM↑ & Id. Acc.↑ \\  
        \midrule
        FaceTTS & \textbf{7.9383} & 0.9048 & 0.6032 & \textbf{7.8128} & 0.8844 & 0.5715  \\ 
        Ours & 8.918 & \textbf{0.9117} &\textbf{ 0.6223} & 8.9822 & \textbf{0.9002} & \textbf{0.5911}  \\
  \bottomrule
\end{tabularx}
\vspace{-1.0em}
    \caption{
    \textbf{Comparisons on image-conditioned text-to-speech.} We compare speech quality and voice-identity coherence against the FaceTTS~\cite{lee2023facetts}.
    }
    \vspace{-1em}
    \label{tab:imagetts}
\end{table}

\noindent\textbf{Baselines.}
We conduct a comparative analysis against state-of-the-art across two tasks: (1) \textit{Audio-driven video generation}: We compare our model against AniPortrait~\cite{wei2024aniportrait}, Echomimic~\cite{chen2025echomimic}, and Hallo3~\cite{chen2025echomimic}. (2) \textit{Image-conditioned text-to-speech}: We compare our model against FaceTTS~\cite{lee2023facetts}.

\begin{figure*}[t!]
    \centering
    \vspace{-1.5em}
\includegraphics[width=0.97\linewidth, trim={0 0 0 0}, clip]{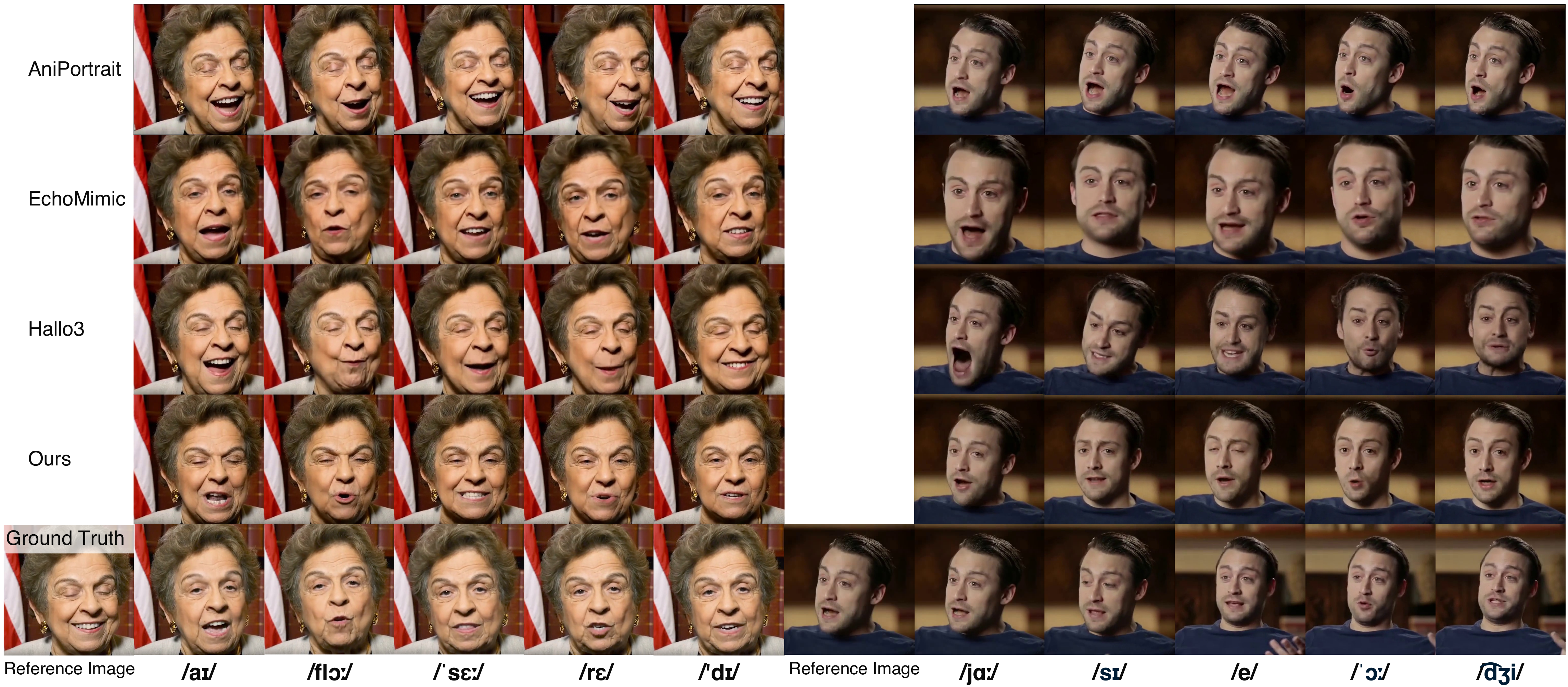}
    \vspace{-1em}
    \caption{\textbf{Comparison on the Speech-driven Video Generation}. We qualitatively compare the video quality and video-audio synchronization against AniPortrait~\cite{wei2024aniportrait}, Echomimic~\cite{chen2025echomimic}, Hallo3~\cite{cui2024hallo3}.}
    \vspace{-1em}
    \label{fig:audio_driven_gen}
\end{figure*}

\noindent\textbf{Metrics.}
We employ a suite of metrics targeting video quality, lip synchronization, and audio fidelity.
We measure the perceptual distributional distance between generated and ground truth videos using Fréchet Inception Distance (FID)~\cite{fid} and Fréchet Video Distance (FVD)~\cite{fvd}, and assess video quality using Q-Align VLM (IQA)~\cite{qalign}.
We quantify lip-sync accuracy using the confidence (Sync-C) and distance (Sync-D) scores from the pre-trained SyncNet model~\cite{syncnet}.
For audio fidelity, we use Mel Cepstral Distortion with Dynamic Time Warping (MCD-DTW)~\cite{mcddtw} to measure the audio quality, and cosine similarity (C-SIM)~\cite{xvector, TE2E} to measure the voice identity coherence.

\noindent\textbf{Speech-driven video generation.}
We provide quantitative comparisons in Tab.~\ref{tab:audio_driven_gen}.
Our model surpasses all baselines on FID and FVD, indicating superior video quality and alignment with the ground truth.
Moreover, our model yields lip synchronization and IQA scores that are comparable to established strong baselines, including Echomimic~\cite{chen2025echomimic} and Hallo3~\cite{cui2024hallo3} that are specifically trained on the benchmark datasets.
This notable performance highlights the exceptional quality of our lip synchronization and overall realism.
We further provide qualitative comparisons in Fig.~\ref{fig:audio_driven_gen}. Notably, given an image of a woman with closed eyes (Fig.~\ref{fig:audio_driven_gen} left), our method and Echomimic successfully synthesize the woman opening her eyes and speaking, whereas AniPortrait~\cite{wei2024aniportrait} and Hallo3 fail to achieve this. We also observe that Hallo3 tends to synthesize exaggerated expressions on accented notes (Fig.~\ref{tab:audio_driven_gen} right). While this boosts Syn-C score, the resulting expressions often appear unnatural. In contrast, our method not only synthesizes a high-quality video with a consistent identity, but also generates realistic lip motions highly synchronized to the accompanying speech.

\noindent \textbf{Image-conditioned speech generation.}
Quantitative results are detailed in Tab.~\ref{tab:imagetts}. Our model achieves superior performance on C-SIM and Identity Accuracy across both datasets, confirming better identity coherence and stronger semantic matching between the reference image and the generated audio identity.
This enhanced identity-to-audio mapping is a direct result of our multimodal training strategy.
However, we observe a slightly lower MCD-DTW score, primarily because our model adopts a lightweight, general-purpose detokenizer, while FaceTTS~\cite{lee2023facetts} employs a heavier audio diffusion model tailored to this task.

\subsection{Ablation Studies}
\begin{table}[!t]
    \centering
\tabcolsep 6.8pt
\scriptsize
\begin{tabularx}{\linewidth}{lccccc}
    \toprule
    \multirow{2}{*}{Methods} & \multicolumn{5}{c}{CelebV-HQ~\cite{zhu2022celebv}}\\
     \cmidrule(l){2-6}
        ~ & FID↓ & FVD↓ & Sync-C↑ & Sync-D↓ & IQA↑ \\  
        \midrule
        w/o Unified Model & 7.279 & 170 & 3.209 & 10.143 & 3.695 \\ 
        w/o Thinking & 13.76 & 128.1 & 3.088 & 10.209 & 3.593 \\
        Full Model & \textbf{6.818} & \textbf{93.81} & \textbf{5.210} & \textbf{8.998} & \textbf{3.794}\\
        \midrule
     \multirow{2}{*}{Methods} & \multicolumn{5}{c}{HDTF~\cite{hdtf}} \\
          \cmidrule(l){2-6}
    ~ & FID↓ & FVD↓ & Sync-C↑ & Sync-D↓ & IQA↑  \\  
        \midrule
        w/o Unified Model & 6.353 & 199.5 & 3.991 & 9.97 & 3.892 \\ 
        w/o Thinking & 13.43 & 110.3 & 4.478 & 9.597 & 3.809  \\
        Full Model &  \textbf{5.779} & \textbf{81.64} & \textbf{6.198} & \textbf{8.822} & \textbf{3.94}\\
  \bottomrule
\end{tabularx}
\vspace{-1.0em}

    \caption{\textbf{Ablation on Design Choices}. We show ablation studies of different designs on speech-driven video generation task.}
    \vspace{-1em}
    \label{tab:ablation}
\end{table}

\noindent\textbf{Unified vs. modality-specific modeling.}
To examine the capacity efficiency of unified multimodal modeling, we compare our model against a baseline composed of multiple expert models, with each producing a single modality trained with identical data, architecture, and training settings. The total parameter count across all experts is set to match that of our unified model.
As shown in Tab.~\ref{tab:ablation}, the unified model outperforms the ensemble of experts across all metrics. This indicates that jointly learning across modalities within a shared architecture yields stronger representations than training isolated experts models.

\noindent\textbf{Effect of thinking-in-modality.} We perform quantitative and qualitative ablations on our modality thinking strategy. As shown in Tab.~\ref{tab:ablation}, we compare two variants for speech-driven video generation. \textit{W/o Thinking} directly generates $\{d_\textrm{sph}, d_\textrm{img}\} \rightarrow [ d_\textrm{vid} ]$, whereas \textit{Full Model} utilizes the full chain of generation: $\{d_\textrm{sph}, d_\textrm{img}\} \rightarrow [ d_\textrm{shp}, d_\textrm{exp}, d_\textrm{sem},  d_\textrm{dsc}, d_\textrm{vid} ]$. Our \textit{Full Model} surpasses the baseline on all metrics, demonstrating that incorporating 3DMM and description as intermediate representations stabilizes video quality and audio synchronization.

\vspace{-0.5em}
\section{Conclusion}
\vspace{-0.5em}
We present Archon, a unified multimodal model for holistic digital human generation.
Archon supports any-modality-to-any generation, understanding, and editing across description, script, speech, animation, semantic video, image, and video.
Archon is a step toward truly unified digital human modeling, where diverse modalities coexist and interact coherently within a single generative framework. Please refer to supplementary Sec.~\textcolor{cvprblue}{B} for more discussions.

{
    \small
    \bibliographystyle{ieeenat_fullname}
    \bibliography{main}
}

\clearpage

\appendix

\renewcommand\thesection{\Alph{section}}
\renewcommand\thetable{\Alph{table}}
\renewcommand\thefigure{\Alph{figure}}

\begin{strip}
\begin{center}
{\huge \bf Supplementary Material}
\end{center}
\end{strip}

\maketitle

This supplementary material provides additional implementation details, experimental results and a brief discussion on ethics. In Sec.~\ref{sec:ethics}, we claim that our work adheres to ethical guidelines. In Sec.~\ref{sec:implement}, we elaborate on the architectural designs and training protocols for our semantic and animation tokenizers, followed by the detailed specifications of our multimodal task formulation. Sec.~\ref{sec:multimodal_data} describes the data acquisition and preprocessing pipelines employed for each modality, including description, script, speech, animation, semantic video and video. Finally, Sec.~\ref{sec:more_results} presents extended qualitative results showcasing multimodal generation and editing capabilities, alongside further quantitative ablation studies on the semantic-guided video diffusion model.

\section{Ethical Considerations}
\label{sec:ethics}
We acknowledge the dual-use nature of high-fidelity avatar generation. While promising for telepresence and content creation, this technology carries risks. We strictly condemn the misuse of our work for harassment or misinformation and emphasize that this research is intended solely for academic purposes. We utilize data in strict adherence to their licenses. We are fully committed to the Ethics Guidelines, advocating for safeguards like invisible watermarking and continuous bias monitoring to ensure responsible deployment.

\section{Discussion}
\label{sec:discussion}
\begin{figure*}[!t]
\centering
\includegraphics[width=1.0\linewidth, trim={0 0 0 0}, clip]{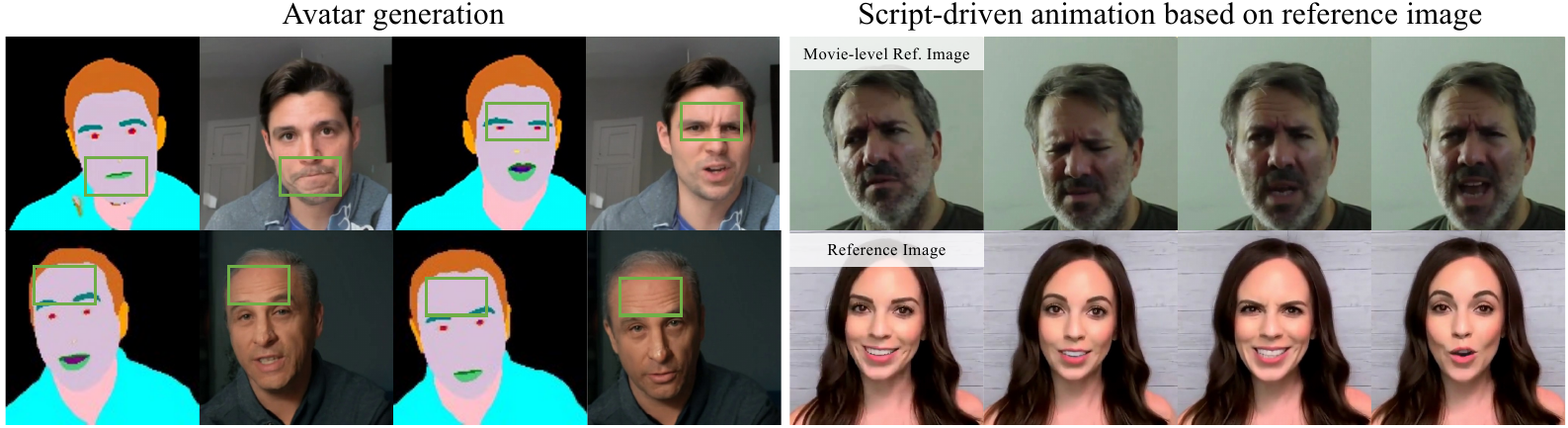}
\caption{\textbf{Micro-expression and rich motion generation.} Qualitative results are presented for video generation from textual inputs and script-driven image animation.}
\label{fig:seg_reliance}
\end{figure*}

Archon employs semantic video as a memory-efficient intermediate representation to compress redundant visual information from RGB frames for cross-modal reasoning. To reconstruct high-quality video output, a semantic-driven diffusion model decodes the semantic sequence under the conditioning of a textual description. This design enables the synthesis of vivid, fine-grained facial dynamics, such as frowning and forehead wrinkles, that are not explicitly encoded in the semantic masks, as illustrated in the left examples of Fig. \ref{fig:seg_reliance}, while also supporting rich, film-grade emotional expression as shown on the right. Moving forward, the semantic representation could be substituted with more detailed, memory-efficient alternatives to capture subtler visual cues. Furthermore, the expressiveness and diversity of avatars may be enhanced by fine-tuning the model on high-quality cinematic datasets.
Additionally, while Archon currently generates single‑speaker talking videos, multi‑turn dialogues, where only one person appears per frame and different persons speak one-by-one, could be realized by integrating an external agent system that parses conversation turns and iteratively queries the model with alternating speaker identities for each turn. Generating multiple persons within the same frame simultaneously, however, would require fine-tuning on dedicated multi‑person datasets.

\section{Implementation Details}
\label{sec:implement}

\begin{table*}[!t]
\centering

\begin{tabular}{@{}lccc@{}}
\toprule
\textbf{Layer} & \textbf{Kernel / Stride} & \textbf{Output Shape} & \textbf{Output Channels} \\ \midrule
Input & - & $T \times H \times W$ & $C_{in}$ \\
\midrule
Conv3D (Strided) & (3, 3, 3) / (1, 2, 2) & $T \times \frac{H}{2} \times \frac{W}{2}$ & $F$ \\
\midrule
\multicolumn{4}{@{}l}{\textbf{For} $i=0$ \textbf{to} $N_{blocks}-1$: (Downsampling Stage $i$)} \\
\quad $N_{res} \times$ ResBlock & (3, 3, 3) / (1, 1, 1) & $T_i \times H_i \times W_i$ & $F \cdot M_i$ \\
\quad Downsample Conv3D & (4, 4, 4) / ($S_{t,i}$, 2, 2) & $T_{i+1} \times H_{i+1} \times W_{i+1}$ & $F \cdot M_{i+1}$ \\
\midrule
$N_{res} \times$ ResBlock & (3, 3, 3) / (1, 1, 1) & $T' \times H' \times W'$ & $F \cdot M_{last}$ \\
\midrule
Group Norm + SiLU & - & $T' \times H' \times W'$ & $F \cdot M_{last}$ \\
Conv3D (to Embedding) & (1, 1, 1) / (1, 1, 1) & $T' \times H' \times W'$ & $D_{emb}$ \\
\bottomrule
\end{tabular}
\caption{The architecture of our 3D CNN Encoder. The input video has dimensions $T \times H \times W \times C_{in}$. $F$ is the base number of filters, $M_i$ are the channel multipliers for each block, $N_{res}$ is the number of residual blocks per stage, and $S_{t,i}$ is the temporal stride for the $i$-th downsampling layer.}
\label{tab:encoder_arch}
\end{table*}

\begin{table*}[h!]
\centering

\begin{tabular}{@{}lccc@{}}
\toprule
\textbf{Layer} & \textbf{Kernel / Stride} & \textbf{Output Shape} & \textbf{Output Channels} \\ \midrule
Input (Quantized Latent) & - & $T' \times H' \times W'$ & $D_{emb}$ \\
\midrule
Conv3D & (3, 3, 3) / (1, 1, 1) & $T' \times H' \times W'$ & $F \cdot M_{last}$ \\
$N_{res} \times$ ResBlock & (3, 3, 3) / (1, 1, 1) & $T' \times H' \times W'$ & $F \cdot M_{last}$ \\
\midrule
\multicolumn{4}{@{}l}{\textbf{For} $i=N_{blocks}-2$ \textbf{down to} $0$: (Upsampling Stage $i$)} \\
\quad $N_{res} \times$ ResBlock & (3, 3, 3) / (1, 1, 1) & $T_{i+1} \times H_{i+1} \times W_{i+1}$ & $F \cdot M_{i+1}$ \\
\quad Upsample + Conv3D & (3, 3, 3) / (1, 1, 1) & $T_i \times H_i \times W_i$ & $F \cdot M_i$ \\
\midrule
Group Norm + SiLU & - & $T \times \frac{H}{2} \times \frac{W}{2}$ & $F$ \\
\midrule
Upsample + Conv3D & (3, 3, 3) / (1, 1, 1) & $T \times H \times W$ & $F$ \\
Conv3D (to Output) & (3, 3, 3) / (1, 1, 1) & $T \times H \times W$ & $C_{out}$ \\
\bottomrule
\end{tabular}
    \caption{The architecture of our 3D CNN Decoder. The input is the latent tensor of shape $T' \times H' \times W' \times D_{emb}$.}
    \label{tab:decoder_arch}
\end{table*}

\subsection{Semantic Video Tokenizer}
\noindent\textbf{Model Architecture.}
Our semantic tokenizer is built upon a 3D convolutional encoder-decoder architecture, designed to map a semantic video to discrete latent codes. The architecture is fully convolutional, consisting of an encoder, a decoder, and look-up free quantizer. We initialize our model's weights from a pre-trained MAGVIT-v2~\cite{yu2024language} checkpoint and introduce a key architectural modification to achieve a higher spatial compression rate suitable for semantic tokenization. Specifically, we add an additional downsampling operation at the beginning of the encoder. This is implemented as a 3D convolution with a spatial stride of 2, which immediately halves the spatial resolution ($H \times W$) of the input video. Correspondingly, a final upsampling block is added at the end of the decoder to restore the original resolution. This modification doubles the spatial downsampling factor of the feature maps before they are quantized, leading to a more compressed grid of semantic tokens.
The detailed architectures of the encoder and decoder are presented in Table~\ref{tab:encoder_arch} and Table~\ref{tab:decoder_arch}, respectively.

The encoder processes an input video tensor of shape $T \times H \times W \times C_{in}$. It begins with a $3 \times 3 \times 3$ 3D convolution with a stride of $(1, 2, 2)$, halving the spatial dimensions ($H, W$) from the outset. This is followed by a series of downsampling stages. Each stage consists of multiple residual blocks (ResBlocks) followed by a downsampling layer, which is a strided 3D convolution that reduces spatial resolution and, in some stages, temporal resolution. The number of feature channels is progressively increased through the encoder. The final stage consists of additional ResBlocks and a $1 \times 1 \times 1$ convolution to project the features into the desired embedding dimension, $D_{emb}$, before quantization.

The decoder is architecturally symmetric to the encoder. It takes the quantized latent tensor of shape $T' \times H' \times W' \times D_{emb}$ and reconstructs the video to its original dimensions. It begins with a $3 \times 3 \times 3$ convolution and several ResBlocks. Subsequently, a series of upsampling stages, each comprising multiple ResBlocks and an upsampling layer, progressively increase the spatial and temporal resolution while decreasing the number of channels. Upsampling is performed using nearest-neighbor interpolation followed by a $3 \times 3 \times 3$ convolution. To mirror the encoder's design, the final layer of the decoder is an upsampling block that doubles the spatial resolution, followed by a final convolution to produce the output video with $C_{out}$ channels.

\vspace{2mm}
\noindent\textbf{Training.}
The model is trained as a VQ-GAN~\cite{esser2021taming}, fine-tuning from the aforementioned MAGVIT-v2 checkpoint. The training was conducted on 4 TPU v4 platform and took 140 hours to complete. We train on clips at a $128 \times 128$ spatial resolution, with a batch size of 8. The training objective is a composite loss function designed to produce high-fidelity reconstructions that also align with ground-truth semantic segmentation maps. The overall loss function $\mathcal{L}$ is defined as:
\begin{equation}
\mathcal{L} = \mathcal{L}_{\text{recon}} + \lambda_{\text{adv}}\mathcal{L}_{\text{adv}} + \mathcal{L}_{\text{commit}} + \lambda_{\text{seg}}\mathcal{L}_{\text{seg}}
\end{equation}
where the components are as follows:
$\mathcal{L}_{\text{recon}}$ is L2 reconstruction loss between the original semantic video $x$ and the reconstructed semantic video $\hat{x}$.
$\mathcal{L}_{\text{adv}}$ is the adversarial loss from a patch-based temporal discriminator that encourages perceptual realism. We set the adversarial loss weight $\lambda_{\text{adv}}$ to 0.3.
$\mathcal{L}_{\text{commit}}$ is the commitment loss from the vector quantization layer, which regularizes the latent embedding space.
$\mathcal{L}_{\text{seg}}$ is a pixel-wise cross-entropy loss between the reconstructed output and the ground-truth segmentation map. The logits for this loss are derived from the negative squared L2 distance between the generated pixel colors and a predefined 21-class color palette, sharpened by a temperature $\tau=10$. This loss component is crucial for learning a semantic representation, and we set its weight $\lambda_{\text{seg}}$ to 3.0.
The model is trained end-to-end using the Adam optimizer with a cosine learning rate schedule and a warm-up phase.

\subsection{3DMM Tokenizer}
We develop three distinct tokenizers for identity, expression, and pose respectively, which are fundamental components of a 3D facial parametric model. Each tokenizer is an autoencoder trained using a vector quantization objective (VQ-VAE~\cite{van2017vqvae}). While they operate on different input features, they share the same underlying network architecture. Below, we detail this architecture and the training procedure for each tokenizer.

\vspace{2mm}
\noindent \textbf{Model Architecture.}
Our animation tokenizer is a fully convolutional autoencoder that operates on 1D temporal sequences of animation parameters. The architecture consists of an encoder, a decoder, and a Residual Vector Quantizer (RVQ)~\cite{gray1984vector}. The encoder maps the input sequence to a compressed latent representation, which is then quantized by the RVQ. The decoder reconstructs the original sequence from the quantized latents.
The encoder and decoder are symmetric and built upon a series of 1D convolutional layers and residual blocks (ResBlocks). The specific configuration is detailed in Table~\ref{tab:tokenizer_arch}. The architecture is consistent across all three tokenizers, with the only variation being the input and output feature dimensions.
\begin{table*}[h]
\centering

\begin{tabular}{l|l|l}
\hline
\textbf{Layer} & \textbf{Output Shape} & \textbf{Details} \\
\hline
\multicolumn{3}{c}{\textbf{Encoder}} \\
\hline
Input & $T \times C_{in}$ & $C_{in}$ \\
Conv1D & $T \times F$ & kernel=3, stride=1 \\
Stage 1 & $T/2 \times F$ & 2 $\times$ ResBlock($F$), Downsample \\
Stage 2 & $T/4 \times F$ & 2 $\times$ ResBlock($F$), Downsample \\
Stage 3 & $T/4 \times F$ & 2 $\times$ ResBlock($F$) \\
Bottleneck & $T/4 \times F$ & 2 $\times$ ResBlock($F$) \\
Conv1D & $T/4 \times D$ & kernel=1, stride=1 \\
\hline
\multicolumn{3}{c}{\textbf{Residual Vector Quantizer}} \\
\hline
Quantization & $T/4 \times D$ & Residual VQ \\
\hline
\multicolumn{3}{c}{\textbf{Decoder}} \\
\hline
Input & $T/4 \times D$ & Quantized Latents \\
Conv1D & $T/4 \times F$ & kernel=3, stride=1 \\
Bottleneck & $T/4 \times F$ & 2 $\times$ ResBlock($F$) \\
Stage 1 & $T/2 \times F$ & 2 $\times$ ResBlock($F$), Upsample \\
Stage 2 & $T \times F$ & 2 $\times$ ResBlock($F$), Upsample \\
Stage 3 & $T \times F$ & 2 $\times$ ResBlock($F$) \\
Conv1D & $T \times C_{out}$ & kernel=3, stride=1 \\
Output & $T \times C_{out}$ & Tanh activation \\
\hline
\end{tabular}
\caption{Architecture of the Animation Tokenizer. We denote the input sequence length as $T$ and the feature dimension as $C_{in/out}$. The number of residual blocks per stage is set to 2. The base number of channels is $F=1024$, and the latent dimension is $D=128$. Downsampling and upsampling operations have a stride of 2.}
\label{tab:tokenizer_arch}
\end{table*}

\noindent\textbf{Training.}
The three tokenizers for identity, expression, and pose are trained independently on their respective data streams. 
First, each set of animation parameters (expression, identity, and pose) is normalized independently. We compute the mean and standard deviation for each parameter dimension across the entire training dataset. Then, for each sample, we subtract the mean and divide by the standard deviation. This standardizes the distribution of each parameter, which is essential for stable training of the subsequent quantization model.
Second, the training objective is to minimize the reconstruction error, regularized by the vector quantization loss. The total loss function is a weighted sum of an L2 reconstruction loss and a VQ commitment loss, as defined in the original VQ-VAE work.
All models are trained using the AdamW optimizer with a cosine learning rate schedule, including a warm-up phase over the first 0.5\% of training steps. The base learning rate is set to $4 \times 10^{-4}$. The training is performed using 128 TPU v2. The specific hyperparameters for each tokenizer are detailed below:
\begin{itemize}
    \item Identity Tokenizer: Trained for 200k steps with a global batch size of 1024. The RVQ consists of 12 codebooks, each with a size of 512. The reconstruction loss weight is 50.0, and the commitment loss weight is 1.0.
    \item Expression Tokenizer: Trained for 200k steps with a global batch size of 1024. The RVQ uses 8 codebooks, each of size 2048. The loss weights are the same as for the identity tokenizer.
    \item Pose Tokenizer: Trained for 100k steps with a larger global batch size of 1024 to stabilize training. The RVQ has 12 codebooks, each with a size of 512. The loss weights are the same as for the identity tokenizer.
\end{itemize}

\begin{table*}[!t]
\centering

\begin{tabular}{@{} l p{0.8\textwidth} @{}}
\toprule
\textbf{Output Modality} & \textbf{Input Modality Combinations} \\
\midrule
\texttt{script (c)} & \texttt{[desc (t)]}, \texttt{[script (p)]}, \texttt{[speech (c)]}, \texttt{[expr (c)]}, \texttt{[semantic (c)]} \\
\midrule
\texttt{speech (c)} & \texttt{[desc (t)]}, \texttt{[desc (t), script (c)]}, \texttt{[script (c)]}, \texttt{[script (c), speech (p)]}, \texttt{[speech (p)]}, \texttt{[script (c), image (t)]}, \texttt{[script (c), semantic (c)]}, \texttt{[script (c), speech (p), semantic (c)]}, \texttt{[id (t), expr (c)]}, \texttt{[semantic (c)]}, \texttt{[script (c), id (t)]} \\
\midrule
\texttt{image (t)} & \texttt{[desc (t)]}, \texttt{[speech (c)]}, \texttt{[id (t)]}, \texttt{[id (t), expr (t), pose (t)]}, \texttt{[desc (t), id (t), expr (t), pose (t)]} \\
\midrule
\texttt{identity (t)} & \texttt{[desc (t)]}, \texttt{[desc (t), script (c), speech (c)]}, \texttt{[speech (c)]}, \texttt{[image (t)]} \\
\midrule
\texttt{expression (c)} & \texttt{[desc (t)]}, \texttt{[desc (t), script (c), speech (c), image (t), id (t)]}, \texttt{[script (c)]}, \texttt{[script (c), speech (c), id (t)]}, \texttt{[speech (c)]}, \texttt{[speech (c), id (t)]}, \texttt{[speech (c), id (t), image (t)]}, \texttt{[speech (c), image (t)]}, \texttt{[semantic (c)]}, \texttt{[expr (p), speech (c)]} \\
\midrule
\texttt{pose (c)} & \texttt{[desc (t)]}, \texttt{[speech (c)]}, \texttt{[speech (c), id (t), expr (c)]}, \texttt{[speech (c), expr (c), image (t)]}, \texttt{[speech (c), id (t), expr (c), image (t)]}, \texttt{[image (t), id (t), expr (c)]}, \texttt{[image (t), expr (c)]}, \texttt{[id (t), expr (c)]}, \texttt{[semantic (c)]} \\
\midrule
\texttt{semantic (c)} & \texttt{[desc (t)]}, \texttt{[desc (t), script (c)]}, \texttt{[desc (t), script (c), speech (c)]}, \texttt{[desc (t), script (c), speech (c), id (t), expr (c), pose (c)]}, \texttt{[desc (t), image (t)]}, \texttt{[desc (t), script (c), speech (c), id (t), expr (c), pose (c), image (t)]}, \texttt{[script (c)]}, \texttt{[script (c), image (t)]}, \texttt{[script (c), speech (c), id (t), expr (c), pose (c), image (t)]}, \texttt{[speech (c)]}, \texttt{[speech (c), id (t), expr (c), pose (c)]}, \texttt{[speech (c), id (t), expr (c), pose (c), image (t)]}, \texttt{[speech (c), image (t)]}, \texttt{[speech (c), expr (c), pose (c), image (t)]}, \texttt{[image (t)]}, \texttt{[image (t), id (t), expr (c), pose (c)]}, \texttt{[image (t), expr (c), pose (c)]}, \texttt{[image (t), expr (c)]}, \texttt{[id (t), expr (c), pose (c)]}, \texttt{[semantic (p)]}, \texttt{[semantic (p), speech (c)]}, \texttt{[semantic (p), expr (c), pose (c)]} \\
\midrule
\texttt{description (t)} & \texttt{[speech (c)]}, \texttt{[image (t), semantic (c), speech (c)]}, \texttt{[image (t), semantic (c), speech (c), script (c)]}, \texttt{[image (t)]}, \texttt{[image (t), semantic (c)]}, \texttt{[id (t), expr (c), pose (c)]} \\
\bottomrule
\end{tabular}
\caption{Overview of the 72 Multimodal Training Tasks. The model is trained to predict the output modality from various input combinations. The state of each modality is indicated in parentheses: (c) for current, (p) for past, and (t) for time-invariant. For brevity in long combinations, we use abbreviations: \texttt{desc} (description), \texttt{id} (identity), and \texttt{expr} (expression).}
\label{tab:multimodal_tasks}
\end{table*}

\subsection{Multimodal Task Formulation and Training}

Our primary objective is to train a single, versatile multimodal model capable of holistic avatar generation. This necessitates that the model understands and generates the wide array of modalities that constitute a digital avatar. To this end, we have meticulously designed a comprehensive suite of 72 multimodal tasks. These tasks are structured to teach the model not only to generate individual modalities but also to understand the intricate relationships and dependencies between them, enabling a cohesive and realistic generation of a holistic avatar.

The model is trained on a rich set of modalities, encompassing textual (\texttt{description}, \texttt{script}), acoustic (\texttt{speech}), semantic (\texttt{semantic}), and a hierarchical set of visual modalities (\texttt{identity}, \texttt{expression}, \texttt{pose}, \texttt{image}). Modalities such as \texttt{identity}, \texttt{image}, and \texttt{description} are considered time-invariant. The other modalities can be either from \textit{past} or \textit{current} depending on if they are the conditions or the predictions in task definition.

Our training tasks are formulated as sequence-to-sequence problems, where the model is given a set of input modalities and is asked to generate a target output modality. The tasks can be categorized as follows:
\begin{itemize}
    \item \textbf{Continuation Tasks:} These tasks involve predicting the \textit{current} state of a modality given its \textit{past} state (e.g., \texttt{speech (past)} $\rightarrow$ \texttt{speech (current)}). This helps the model learn the temporal dynamics of the modality.
    \item \textbf{Cross-Modal Generation Tasks:} The majority of our tasks fall into this category. The model learns to generate a target modality from one or more different source modalities (e.g., \texttt{speech (current)}, \texttt{identity (time-invariant)} $\rightarrow$ \texttt{expression (current)}).
    \item \textbf{Chained Generation Tasks:} The tasks are designed to be composable, enabling a chained generation pipeline in our "Thinking in Modality" (e.g., \texttt{image (time-invariant)} $\rightarrow$ \texttt{identity (time-invariant)}, then \texttt{speech (current)} $\rightarrow$ \texttt{expression (current)}, etc.). Our task suite includes all these intermediate steps to facilitate such chained inference.
\end{itemize}
This extensive set of 72 tasks, detailed in Table~\ref{tab:multimodal_tasks}, ensures that the model is exposed to a vast number of input-output combinations, fostering a deep multimodal understanding and enabling the generation of high-fidelity, holistic avatars.

\section{Multimodal Data Details}
\label{sec:multimodal_data}
In this section, we provide a detailed description of the data acquisition and preprocessing pipeline for the all modalities used in our model: description, script, speech, animation, semantic video and image/video. Our pipeline is designed to extract temporally synchronized multimodal data from a large-scale video source, which are essential for training our holistic avatar generation model.
\subsection{Description}
\label{sec:description_modality}

The description modality provides a rich, structured representation of the person in the video. This modality is designed to capture a holistic set of attributes encompassing appearance, actions, and environmental context. To ensure consistency and comprehensiveness, we employ Gemini 2.5 Pro~\cite{comanici2025gemini} for the annotation process. A detailed prompt, shown in Figure~\ref{fig:gemini_prompt_full}, guides the model to analyze a video and return a structured JSON object containing all discernible attributes.

The resulting JSON object is organized into three primary keys: \texttt{appearance}, \texttt{action}, and \texttt{environment}.

\begin{itemize}
    \item \textbf{\texttt{appearance}}: This category captures the subject's physical characteristics. It includes static attributes such as \texttt{gender}, \texttt{age\_group}, \texttt{ethnicity}, and \texttt{body\_build}, as well as more detailed features like \texttt{hair\_color}, \texttt{hair\_style}, \texttt{facial\_features}, and detailed descriptions of \texttt{clothing} and other \texttt{physical\_attributes}.
    \item \textbf{\texttt{action}}: This section details the dynamic aspects of the subject's performance. It describes the overall \texttt{activity\_type}, emotional \texttt{expression}, and nuanced behaviors such as \texttt{mouth\_action}, \texttt{eyebrow\_action}, \texttt{head\_action}, and \texttt{gaze\_direction}. It also captures the overall \texttt{emotion} and \texttt{energy\_level} conveyed.
    \item \textbf{\texttt{environment}}: This category provides context for the scene, describing the \texttt{lighting\_conditions}, \texttt{background\_description}, and the general \texttt{scene\_context} (e.g., indoor/outdoor).
\end{itemize}

This structured approach ensures that our model is trained on a consistent and detailed descriptive modality, enabling it to generate holistic and high-fidelity avatars. An example of the generated JSON data is presented in Fig.~\ref{fig:json_example}.

\subsection{Script}
The script modality is derived from word-level caption data, which includes precise start timestamps and durations for each word. This allows for fine-grained alignment between the text, speech, and video.
When a audio-video clip is extracted, we query the caption data to find all words whose time intervals overlap with the clip's duration. These selected words are then concatenated in their original order to form the final script that corresponds precisely to the spoken content within that segment. This method ensures that the script is accurately aligned with its corresponding audio and video, which is critical for training our multimodal model.

The aforementioned process applies to our training data where precise word-level annotations are available. For our test sets, specifically CelebV-HQ~\cite{zhu2022celebv} and HDTF~\cite{hdtf}, ground-truth captions are not provided. Therefore, we utilize OpenAI's Whisper model~\cite{radford2023whisper} to transcribe the audio into text. This allows us to evaluate our model's performance on datasets with automatically generated scripts, reflecting a more realistic, in-the-wild scenario.

\subsection{Speech}
The speech modality is processed to be tightly synchronized with the corresponding video segment. The raw audio track is first resampled to a standard 16~kHz sampling rate. To enhance signal quality and improve model robustness to acoustic variations, we apply a denoising model~\cite{zeghidour2021soundstream} to 50\% of the audio samples, selected at random. The remaining 50\% are left unchanged to improve the robustness of model.
To align the audio with the video, we extract an audio segment corresponding to the sampled video clip using its timestamps. If the source audio is shorter than the target duration at the sampled point, we pad the segment with silence to ensure a consistent length across all samples.

\begin{table}[!t]
\centering

\resizebox{\columnwidth}{!}{%
\begin{tabular}{lll}
\toprule
\textbf{Label Name} & \textbf{Label Name} & \textbf{Label Name} \\
\midrule
background & hair & pupil \\
face & brows & iris \\
neck & lashes & sclera \\
ears & beard & clothes \\
upper\_lip & teeth & glasses \\
lower\_lip & tongue & headwear \\
nostrils & other\_in\_the\_mouth & accessory \\
\bottomrule
\end{tabular}%
}
\caption{The 21 semantic labels used for generating semantic segmentation maps.}
\label{tab:labels}
\end{table}

\subsection{Animation}
The animation modality in our model is derived from 3D Morphable Model (3DMM) parameters~\cite{egger20203d}, which provide a rich, structured representation of the human head's geometry and dynamics. These parameters are not directly available in the raw video data and are obtained through a pre-processing step where a 3DMM is fitted to each video frame using an optimization-based approach~\cite{danvevcek2022emoca}. This fitting process yields a set of continuous parameters for each frame, which contains facial expression, person-specific identity, head rotation , and head translation.

To make these continuous parameters suitable for our multimodal language model, we convert them into a sequence of discrete tokens. This tokenization process is crucial as it allows our model to handle animation in a manner analogous to text. The process involves two main steps: normalization and vector quantization. First, we use pre-calculated mean and standard deviation of each parameter to normalize itself independently.
Second, the normalized continuous parameters are quantized into discrete tokens using three separate, learned Vector Quantized-Variational Autoencoders~\cite{van2017vqvae} (VQ-VAEs). We use distinct VQ-VAEs for expression, identity, and pose (where pose is the concatenation of rotation and translation parameters). Each VQ-VAE has its own learned codebook, effectively creating a unique vocabulary for each animation component. This process maps the high-dimensional, continuous animation signals into compact sequences of discrete tokens. By tokenizing the animation parameters, we can seamlessly integrate them into our transformer-based MLLM architecture, enabling it to learn cross-modal associations between animation, script, speech, and video frames.

\subsection{Semantic Video}
For the semantic video modality, we process each frame to generate a semantic segmentation map. This provides our model with a context-efficient representation of video, particularly focusing on the human subject.
We utilize the DINOv2 model~\cite{oquab2023dinov2} to parse segmentation, which delineates 21 distinct classes corresponding to key facial features, hair, and accessories. This fine-grained labeling scheme enables our model to learn detailed representations of the human head and upper body. The complete list of the 21 semantic labels is provided in Table~\ref{tab:labels}. This detailed decomposition is instrumental for the model to generate realistic textures and geometry for each part of the avatar, from the subtle nuances of eye components to various accessories.

\subsection{Image/Video}
Our video processing pipeline is designed to generate normalized, head-centric video clips. Initially, raw videos are decoded and uniformly downsampled to a frame rate of 30 frames per second (fps). To achieve a consistent, head-centric view, we first detect the facial landmarks to the face in each frame. From the 3DMM fit, we derive a bounding box that encompasses the face and a portion of the upper body.
To ensure a stable view without jitter, we compute a common crop region based on the union of these bounding boxes across the entire video sequence. The size of this crop is randomly scaled to introduce variations in framing, from tight close-ups to wider shoulder-level shots. This cropped region is then resized to a final resolution of $256 \times 256$ pixels using bicubic interpolation with anti-aliasing.

\begin{table}[!t]
    \centering
\begin{tabular}{lcc}
    \toprule
        Methods/Metrics & FID↓ & FVD↓  \\  
        \midrule
        Ours w/o text cond & 17.2310 & 32.3772 \\ 
        Ours w/o joint crossattention & 22.7003 & 36.1064 \\ 
        Ours & \textbf{16.8678} & \textbf{26.6860} \\ 
  \bottomrule
\end{tabular}
    \caption{
We show the ablation study on the different condition ways to the diffusion model.
    }
    \label{tab:diffusion_ablation}
\end{table}

\begin{figure*}[t!]
    \centering
    \includegraphics[width=0.7\linewidth, trim={0 0 0 0}, clip]{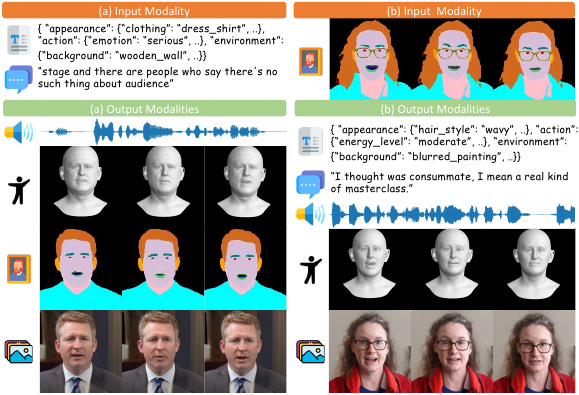}
    \vspace{-0.5em}
    \caption{\textbf{Multimodal Generation.} We show our model is capable of doing text or speech-to-all generation and can support a variety of cross-modality reasoning and generation tasks.}
    \label{fig:video_gen_supp}
\end{figure*}

\section{More Results}
\label{sec:more_results}
\subsection{Multimodal Generation} 
We present additional qualitative results of our multimodal generation framework. As illustrated in Fig.~\ref{fig:video_gen_supp} (a), when conditioned on a description of a subject (e.g., a man in a suit) and a corresponding script, our model successfully synthesizes all complementary modalities, including speech, animation, semantic segmentation, and the final RGB video. These generated modalities exhibit precise temporal synchronization, with the output video demonstrating high visual fidelity and strict alignment with the input conditions. Besides, in Fig.~\ref{fig:video_gen_supp} (b), we demonstrate generation from semantic inputs; given a semantic video, our model synthesizes a photorealistic video that faithfully adheres to the semantic appearance and motion. Furthermore, the model demonstrates its understanding capabilities by accurately inferring the description, script, speech, and animation directly from the semantic video input.

\subsection{Modality-specific Editing}
\begin{figure*}[ht!]
    \centering
    \includegraphics[width=1.0\linewidth, trim={0 0 0 0}, clip]{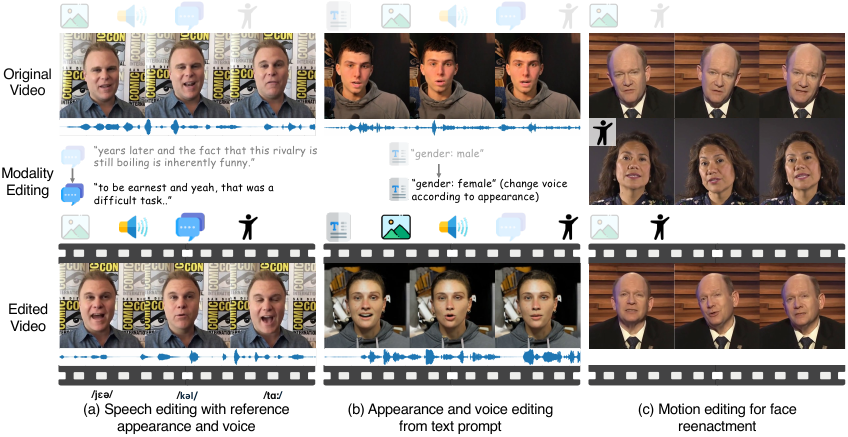}
    \vspace{-1.5em}
    \caption{\textbf{Modality-specific Editing}. We demonstrate that our model support editing mutlimodal input with remarkable flexibility - modifying arbitrary chosen modal while maintains the others untouched. The icons on the top showing the modalities used in the example, and the highlighted icons are the ones that are varied.}
    \label{fig:editing_supp}
\end{figure*}

We further demonstrate our model's versatility in multimodal editing in Fig.~\ref{fig:editing_supp}. In Fig.~\ref{fig:editing_supp} (a), we perform script editing, where we synthesize a new talking video that articulates a modified script while faithfully preserving the subject's original appearance and vocal identity. In Fig.~\ref{fig:editing_supp} (b), we illustrate disentangled attribute editing by altering the speaker's gender from male to female while retaining fine-grained attributes such as hairstyle and clothing. Notably, the model simultaneously adapts the generated voice to align with the modified visual appearance, demonstrating robust cross-modal consistency. Finally, Fig.~\ref{fig:editing_supp} (c) depicts animation editing (face reenactment); by utilizing 3DMM coefficients extracted from a reference video, we drive the source subject to replicate the reference's pose and expressions with high fidelity.

\subsection{Comparisons with Unified Models.}
\label{ssec:compare_unified_models}
Current unified models are not truly ``holistic'' in the context of digital human. While models such as Qwen-Omni~\cite{Qwen3-Omni} can understand multimodal inputs, their outputs are confined to text and audio. Similarly, Show-o2~\cite{xie2025show} supports text, image, and video but omits audio. NExT-GPT~\cite{wu2024next}, though capable of audio generation, produces environmental or musical sounds rather than intelligible speech. In contrast, Archon is the first model to learn the joint distribution of holistic human-centric modalities. 

\noindent \textbf{Video Understanding.} We evaluate human talking video understanding against Qwen-Omni~\cite{Qwen3-Omni}, a 30B Mixture-of-Experts model. The test set, drawn from HDTF~\cite{hdtf}, is annotated using Gemini3~\cite{gemini3} to provide ground-truth descriptions. The query 2prompt is shown in Fig.~\ref{fig:gemini_prompt_full} and is also used for Qwen-Omni. Performance is measured by the average F1 score across all descriptive attributes, where precision \(P_s\) and recall \(R_s\) are derived from the cosine similarity of BGE embeddings~\cite{bge_m3}:
\[
P_s = \frac{1}{|\mathcal{P}|} \sum_{p \in \mathcal{P}} \max_{g \in \mathcal{G}} \cos\left( \text{BGE}(p), \text{BGE}(g) \right),
\]
\[
R_s = \frac{1}{|\mathcal{G}|} \sum_{g \in \mathcal{G}} \max_{p \in \mathcal{P}} \cos\left( \text{BGE}(g), \text{BGE}(p) \right).
\]
with \(\mathcal{P}\) and \(\mathcal{G}\) denoting the predicted and ground-truth attribute sets, respectively. Despite its compact 1B parameter size, Archon attains an F1 score of 0.90, which is comparable to the 0.93 achieved by the substantially larger Qwen-Omni model.

\begin{figure*}[t!]
\centering
\vspace{-1.0em}
\includegraphics[width=1.0\linewidth, trim={0 0 0 0}, clip]{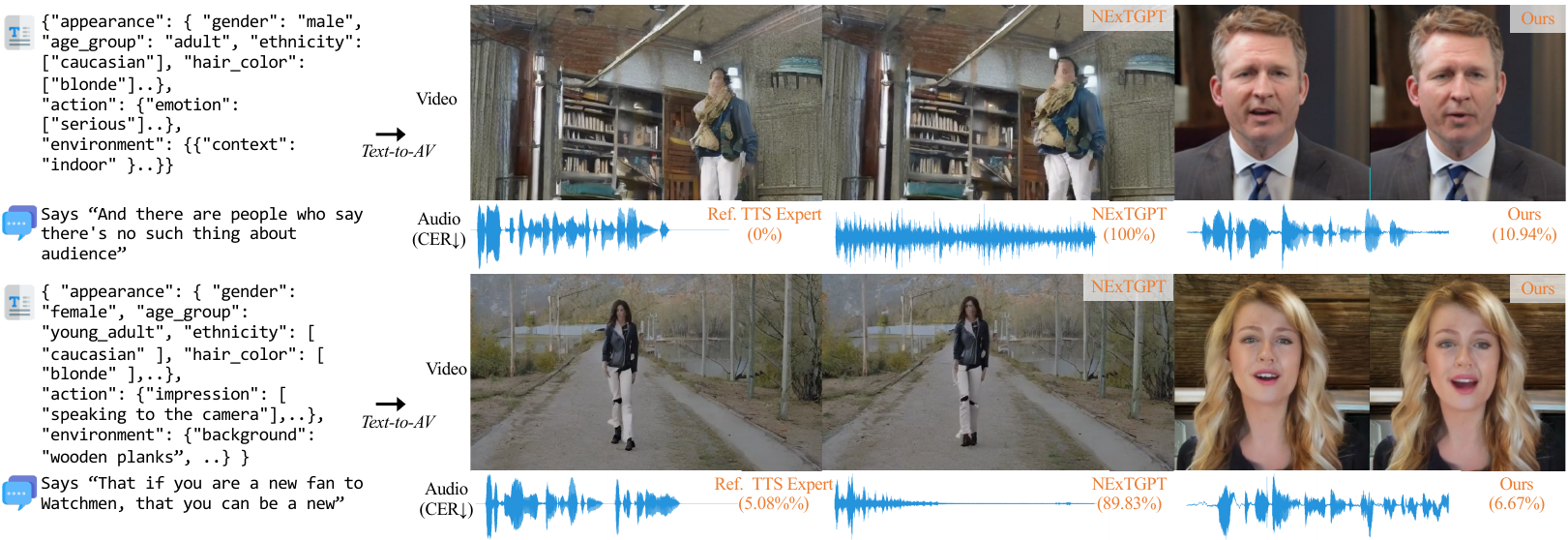}
\caption{\textbf{Text-to-AV Comparisons.} We present a comparative evaluation against NExT-GPT for the generation of audio and video from textual input (comprising both description and script).}
\label{fig:nextgpt_fig}
\end{figure*}

\noindent \textbf{Text-to-AV}. For talking video and audio synthesis from text, we compare with NExT-GPT~\cite{wu2024next}, the most closely related unified model. Input prompts define human attributes and dialogue scripts, as shown on the left of Fig.~\ref{fig:nextgpt_fig}. We use Gemini~\cite{gemini3} to translate this JSON-formatted description to natural language and input it to NExT-GPT. As illustrated in Fig.~\ref{fig:nextgpt_fig}, we synthesize significantly higher-quality video and audio than NExT-GPT. NExT-GPT produces visually corrupted video lacking recognizable facial features or motion, along with unintelligible or silent audio. We generate high-fidelity talking head video that accurately reflects the provided description, with well-synchronized lip movements and clear audio corresponding to the input script.

\subsection{Ablation of Semantic-driven Video Diffusion Model}

We investigate the impact of different conditioning mechanisms on the video diffusion backbone, with quantitative results summarized in Tab.~\ref{tab:diffusion_ablation}. All ablations are performed on the held-out test split. Given the dense temporal nature of the semantic video input, we employ channel-wise concatenation with the noisy latents rather than cross-attention, adhering to computational memory constraints. In Sec.~\textcolor{red}{3.3}, we propose concatenating the reference image with its corresponding segmentation mask for cross-attention (``Ours''). This strategy establishes a structural bridge, facilitating the effective transfer of appearance from the reference to the synthesized video.
We contrast this with a baseline that utilizes only the RGB reference image in the cross-attention module (``Ours w/o joint cross-attention''). As observed in Tab.~\ref{tab:diffusion_ablation}, the degradation in FID and FVD scores in the absence of segmentation image underscores their critical role in guiding the diffusion model to accurately map reference appearance to generated motion. Finally, we assess the contribution of textual prompts. The configuration ``Ours w/o text cond'' omits text embeddings from the cross-attention layers. The resulting decline in metrics confirms that multimodal conditioning is essential for maintaining high video fidelity.

\begin{figure*}[t]
\begin{lstlisting}[language=json,  breaklines=true, basicstyle=\footnotesize\ttfamily]
{
  "appearance": {
    "gender": "male",
    "age_group": "adult",
    "ethnicity": ["caucasian"],
    "body_build": ["average"],
    "hair_color": ["black"],
    "hair_style": ["short"],
    "facial_features": ["none_discernible"],
    "clothing": {
      "upper_body": "button-down shirt over t-shirt",
      "lower_body": "none",
      "footwear": "none",
      "accessories": ["watch", "ring", "bracelet"],
      "dominant_colors": ["black", "blue", "white", "red"]
    },
    "physical_attributes": {
      "visible_tattoos": "none",
      "visible_piercings": "none",
      "distinctive_marks": "none",
      "posture": "upright",
      "gait": "not_applicable",
      "physical_aids": ["none"]
    }
  },
  "action": {
    "activity_type": "speaking",
    "expression": "smile",
    "overall_impression": ["gesturing with hands while speaking", "raising hands to chest level", "making a fist"],
    "emotion": ["positive and engaging", "enthusiastic"],
    "energy_level": ["medium to high", "animated"],
    "mouth_action": ["wide_opening", "narrow_opening", "relaxed_lips", "pulling_of_lip_corners", "synchronization_and_precision_with_spoken_words"],
    "eyebrow_action": ["raising_inner_eyebrows", "raising_outer_eyebrows", "brow_furrowing"],
    "blink_frequency": "medium",
    "head_action": ["mostly_centered", "tilts_inquisitively", "nods_rhythmically", "amplitude_and_tempo_of_movements", "directional_changes_pitch/yaw/roll", "frequency_of_changes", "transitions_between_stillness_and_motion"],
    "eye_state": ["fully_wide_alert"],
    "gaze_direction": ["straight_ahead", "fixed_forward_focus"],
    "nonverbal_habits": ["habitual_eyebrow_flicks", "frequent_micro-nods", "timing_and_context_of_cues", "consistency_of_cues", "typical_amplitude_subtle_vs_pronounced"],
    "interactions": ["with_object_e.g._microphone/instrument", "none"],
    "props_used": ["microphone", "none"]
  },
  "environment": {
    "lighting_conditions": "bright",
    "background_description": "light colored wall with a framed certificate and a model of a watch on a shelf",
    "time_of_day": "unknown",
    "weather_conditions": "indoor_not_applicable",
    "context": "indoor"
  }
}
\end{lstlisting}
\caption{An example of the structured JSON description obtained from Gemini 2.5 Pro.}
\label{fig:json_example}
\end{figure*}

\begin{figure*}
\vspace{-1.0em}
\begin{lstlisting}[
    breaklines=true,
    basicstyle=\scriptsize\ttfamily % You can change this to \scriptsize / /footnotesize if needed
]
Enhanced Video/Image Analysis Prompt
You are an expert video and image analysis AI. Your task is to meticulously analyze a given video or sequence of images featuring a person performing, and extract all possible relevant attributes about the person, their actions, and the surrounding environment.

Critical Instruction for JSON Output
Your output MUST be a single, complete, and perfectly valid JSON object. Do NOT include any introductory or concluding text, explanations, or extraneous characters outside of the JSON. Ensure proper syntax, including all commas, brackets, and quotes.

Data Extraction Schema
Here's the information you need to extract, grouped into three main categories, and the precise JSON format you must adhere to. For attributes specified as lists (e.g., body_build, hair_color, hair_style, facial_features, accessories, dominant_colors, physical_aids, interactions, props_used, overall_impression, Mouth_Action, Eyebrow_Action, Head_Action, Eye_State, Gaze_Direction, Nonverbal_Habits), ensure you return a list containing ALL discernible attributes. The attribute values given below are just examples, but use them for inspiration. Attributes should be strings with undescores instead of space where_necessary.

{
  "appearance": {
    "gender": "male/female/non-binary/unknown",
    "age_group": "child/preteen/teenager/young_adult/adult/older_adult/senior/unknown",
    "ethnicity": ["list_ethnicities_if_discernible", "caucasian", "asian", "african_american", "hispanic", "middle_eastern", "unknown", "etc."],
    "body_build": ["slim", "average", "athletic", "muscular, "heavy", "unknown"],
    "hair_color": ["black", "brown", "blonde", "red", "gray", "white", "green", "blue", "pink", "other", "unknown"],
    "hair_style": ["bald", "long", "short", "curly", "straight", "wavy", ...],
    "eye_color": ["light_brown", "dark_brown", "light_blue", ...],
    "eye_style": ["long_eyelashes", "short_eyelashes", ...],
    "teeth": ["straight", "missing_teeth",...],
    "facial_features": ["glasses", "beard", "wrinkly_skin", "elastic_skin", "clean_skin", "sideburns", "mustache", "freckles", ...],
    "clothing": {
      "upper_body": ["description_of_upper_garment_e.g._t-shirt/dress_shirt/blouse/hoodie/jacket/sweater/tank_top/sports_bra/suit/tie/none"],
      "lower_body": ["description_of_lower_garment_e.g._jeans/trousers/skirt/shorts/leggings/sweatpants/none"],
    },
    "physical_attributes": {
      "visible_tattoos": "description_of_tattoos_and_location_if_visible_e.g._full_sleeve_right_arm/small_design_neck/none",
...
    }
  },
  "action": {
    "activity_type": "dancing/singing/playing_instrument/speaking/acting/sports/walking/running/sitting/standing/other",
    "expression": "smile/frown/closed mouth/etc.",
    "overall_impression": ["brief_descriptive_summary_of_the_performance", "e.g._head_turn_to_the_left", "e.g._raising_her_left_arm"],
    "emotion": ["description_of_general_energy_and_dominant_emotional_tone", "e.g._animated_and_joyful", ...],
    "energy_level": ["overall_magnitude_of_emotional_display", "e.g._faint_micro-expressions_to_vivid_reactions", "e.g._consistently_low-key"],
...
  },
  "environment": {
    "lighting_conditions": "bright/dim/natural_light/artificial_light/backlit/etc.",
    "background_description": "detailed_description_of_the_background_e.g._blurred_trees/crowded_street/empty_white_wall",
    "time_of_day": "morning/afternoon/evening/night/unknown",
    "weather_conditions": "sunny/cloudy/rainy/snowy/indoor_not_applicable/etc.",
    "scene_context": "indoor/outdoor/stage/street/home/office/natural_environment/etc."
  }
}
\end{lstlisting}
\caption{The part of the prompt provided to Gemini 2.5 Pro for video annotation.}
\label{fig:gemini_prompt_full}
\end{figure*}

\end{document}